\documentclass[11pt]{article}

\usepackage[final]{acl}

\usepackage{times}
\usepackage{latexsym}
\usepackage[T1]{fontenc}
\usepackage[utf8]{inputenc}
\IfFileExists{microtype.sty}{\usepackage{microtype}}{}

\usepackage{tabularx}
\usepackage{array}

\usepackage{xspace}
\usepackage{booktabs}
\usepackage{amsmath}
\usepackage{amssymb}
\newif\ifhasenumitem
\IfFileExists{enumitem.sty}{\hasenumitemtrue\usepackage{enumitem}}{\hasenumitemfalse}
\usepackage{graphicx}
\usepackage{xcolor}
\IfFileExists{pifont.sty}{\usepackage{pifont}}{}
\usepackage{url}

\newcommand{\methodshort}{JudgeMoE\xspace}

\newcommand{\best}[1]{\textbf{#1}}
\IfFileExists{pifont.sty}{

}{

}

\ifhasenumitem
  \setlist[itemize]{leftmargin=*, topsep=2pt, itemsep=1pt}
  \setlist[enumerate]{leftmargin=*, topsep=2pt, itemsep=1pt}
\fi

\title{JudgeMoE: Distributional Aggregation for LLM-as-a-Judge}

\author{
\textbf{Yiqi Liu}\textsuperscript{1} \quad
\textbf{Joseph James}\textsuperscript{2} \quad
\textbf{Yang Wang}\textsuperscript{1}
\\
\textbf{Kun Zhao}\textsuperscript{3} \quad
\textbf{Chenghao Xiao}\textsuperscript{4} \quad
\textbf{Chenghua Lin\textsuperscript{1}\thanks{Corresponding author}}
\\[5pt]
\normalfont
\textsuperscript{1}\,The University of Manchester
\quad
\textsuperscript{2}\,The University of Sheffield
\\
\textsuperscript{3}\,University of Pittsburgh
\quad
\textsuperscript{4}\,Shanghai University of Finance and Economics
\\[4pt]
\texttt{\{yiqi.liu, chenghua.lin\}@manchester.ac.uk}\\
}

\begin{document}
\maketitle

\begin{abstract}
When an LLM judge scores an output, its score distribution retains uncertainty and disagreement information that is lost after scalar compression. We introduce JudgeMoE, a lightweight aggregator that assigns example-specific weights to cached judge score distributions and fuses them before computing a final score. A protocol study shows that score-range choice is unstable across judge--dataset settings and that soft scoring usually outperforms hard decoding. On the original 10-cell benchmark, JudgeMoE improves mean Spearman over uniform log pooling by $+0.079$. Applying the same configuration to six additional cells yields a $+0.0393$ mean gain over the strongest local single judge across 16 cells, with positive differences in 12/16 cells and a one-sided Wilcoxon signed-rank $p=0.0091$. Validation-based analyses further show that the preferred aggregation method depends on the task and judge pool.
\end{abstract}

\section{Introduction}

\begin{figure*}[t]
\centering
\includegraphics[width=2.1\columnwidth]{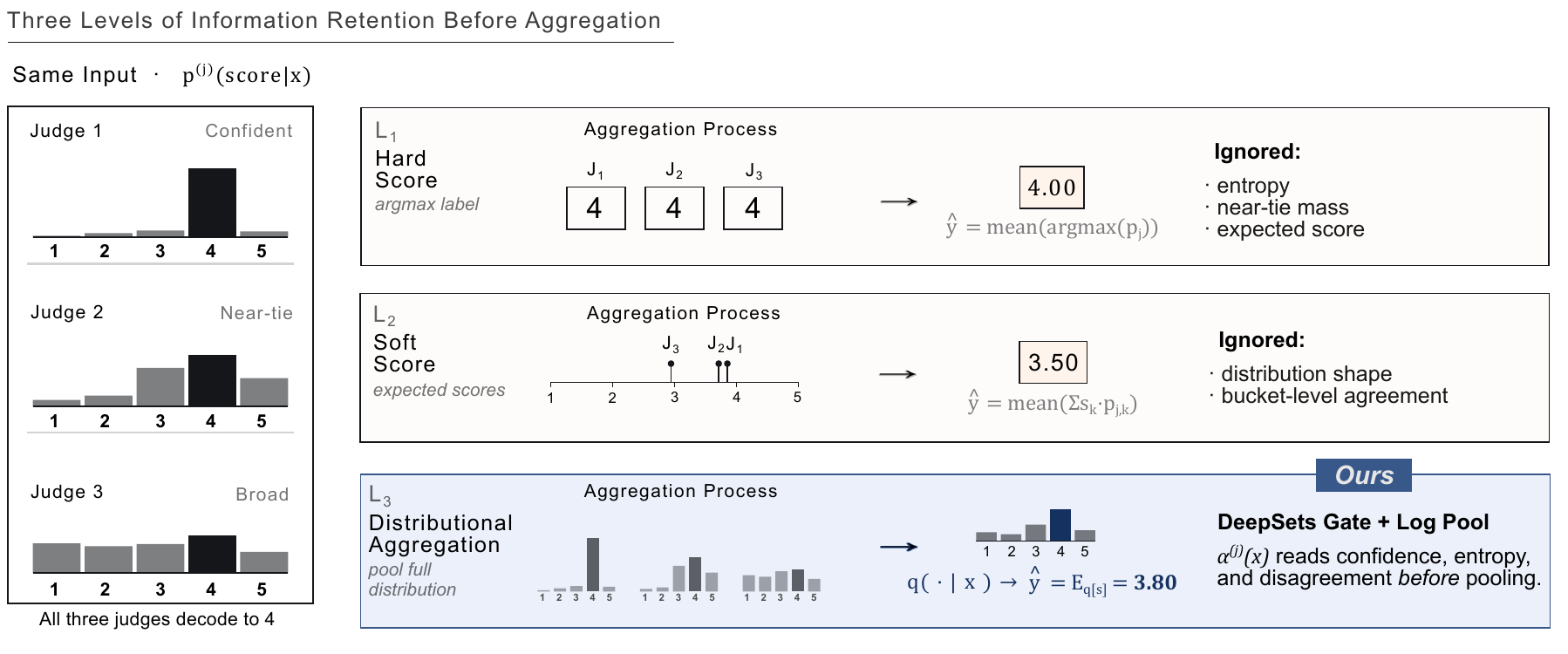}
\caption{Three levels of information retention for the same judge score distributions. Hard-score aggregation uses only decoded labels, and soft-score aggregation uses expected scores. \methodshort pools the full score distributions before scoring, retaining confidence, entropy, near-tie mass, and bucket-level agreement during aggregation.}
\label{fig:judgemoet-overview}
\end{figure*}


When an LLM judge assigns a score, the decoded label is only a compressed view of a richer score distribution. Two judges may both decode to 4 while expressing different uncertainty: one may concentrate mass on 4, while another spreads mass across 3, 4, and 5. Such differences matter for aggregation because confidence, entropy, and near-ties affect whether judges provide consistent or uncertain evidence. We use \emph{scoring protocol} to mean the score range, extraction rule, and prompt framing used to elicit this distribution.

LLMs are now widely used as automatic evaluators for summarization, dialogue, reasoning, and translation \citep{liu2023geval, zheng2023judging}. Their agreement with humans can change under protocol choices such as the score range and the rule used to extract a score \citep{wang2024fair, yamauchi2025designchoices, li2026scaleimpact}. Our protocol study shows that these choices change the measurement produced by the judge: across 30 dataset and model combinations where both soft and hard scores give usable rank correlations, each of the five tested score ranges is strongest in at least one setting, and soft-score extraction is stronger than hard decoding in most valid runs.

These findings motivate aggregation before score compression. Scalar averaging and scalar-gated aggregation combine decoded or expected scores after each judge distribution has already been reduced to one number. \methodshort instead pools the full judge score distributions and computes the final scalar score after pooling. Each judge distribution is treated as protocol-specific evidence over the benchmark's possible labels.

We use \methodshort as a lightweight testbed for this representation shift. With a frozen pool of small local judges and cached score distributions, the same outputs support several validation-guided deployment choices. A DeepSets gate assigns example-specific judge weights, and a tempered log-opinion pool fuses score buckets before final scoring. Our experiments focus on reusable pools of small local judges, while larger evaluators are included as standalone options in the validation-based selection procedure. We call one dataset paired with one score-scale condition a \emph{cell}.


Our contributions are as follows:
\begin{itemize}[noitemsep,topsep=1pt,parsep=1pt,partopsep=1pt]
\item A protocol study showing that score-range choice is unstable across judge families and datasets, and that soft scoring is typically stronger than hard decoding.
\item We introduce \methodshort, a lightweight conditional aggregator that builds on the protocol study by pooling cached judge score distributions before converting them to scalar scores.
\item A validation-guided deployment analysis showing when fixed pooling, full \methodshort, sparse repair, or a standalone expert is supported by validation evidence.
\end{itemize}

\section{Related Work}

\paragraph{LLM-as-a-Judge.}
Recent work has established LLMs as scalable evaluators for open-ended generation \citep{liu2023geval, zheng2023judging}. Later studies show that positional bias and evaluator self-preference can affect agreement with humans \citep{wang2024fair, liu-etal-2024-llms-narcissistic}. Prompt design, safety artifacts, decoding choices, and grading scales can also change evaluation outcomes \citep{hong-etal-2026-beyond, chen2025safer, yamauchi2025designchoices, li2026scaleimpact}. Other work improves individual judges through specialized evaluator models \citep{kim2024prometheus2, gu-etal-2025-finetunedjudge}, self-rationalization \citep{trivedi2024selfrationalization}, checklist decomposition \citep{lee-etal-2025-checkeval}, and adaptive debiasing \citep{yang2026fairjudge}. Much of this line studies how to make a single judge more reliable; our work asks what can still be recovered by aggregating across multiple judges that produce different scores for the same example under a shared scoring protocol.

\paragraph{Evaluator ensembles and aggregation.} Multi-model evaluation can use interaction, aggregation, or probability-level comparison across evaluators. Debate-style approaches let agents exchange rationales before producing a final answer \citep{chan2023chateval,du2023debate}; jury-style approaches aggregate judgments from several models without explicit multi-turn interaction \citep{verga2024poll,zhao2025lmc}; and ContrastScore uses disagreement in model probabilities for generation evaluation \citep{wang-etal-2025-contrastscore}. \citet{wang2025judgmentdistribution} study inference from the judgment distribution of a single LLM evaluator and show that distributional information can improve over greedy decoded judgments. \methodshort studies conditional aggregation across multiple judges, using per-example score distributions under explicit scoring protocols and example-specific judge weights.

There is an established statistical tradition of aggregating noisy judgments under latent reliability models, from Dawid--Skene \citep{dawid1979maximum} to recent LLM-specific probabilistic aggregation models \citep{zhao2025voices}. These models infer consensus labels and judge reliability from noisy annotations. They assume a shared discrete label space and a single latent consensus label, whereas we treat score distributions as protocol-dependent measurements and compare several deployment modes over cached outputs. Relatedly, QAEval uses the phrase ``mixture of evaluators'' for heterogeneous QA evaluators that mix rule-based and learned components \citep{yue-etal-2025-qaeval}; our study focuses on a homogeneous LLM pool and treats heterogeneous experts as optional task-aligned additions.

\paragraph{Adjacent settings.}
Pairwise preference learning provides relative supervision that can complement direct scores \citep{bradley1952rank, jeong2025comparative}. RLHF work also uses reward-model ensembles or judge-based constraints inside policy optimization \citep{zhang2024efficientrmensemble, zhai2024uprlhf, xu2024judgeconstraintsrlhf}. Our setting is complementary: we aggregate frozen LLM judges directly at evaluation time, so one cached set of score distributions can support several validated aggregators without rerunning judge inference. This reuse objective distinguishes our setting from post-training evaluator ensembles.
\section{Method}

\paragraph{Setup and terminology.}
\methodshort takes each judge's probability distribution over possible scores as the input object. We use \emph{bucket} as shorthand for one allowed score label, and a \emph{bucket vector} is the normalized probability vector over those labels. A \emph{protocol} is the score range, prompt, and score-extraction rule that produced that vector. A \emph{cell} is one dataset paired with one score scale. A \emph{gate} assigns an example-specific weight to each judge. A \emph{log pool} fuses the weighted score distributions by taking a weighted geometric mean over buckets. \emph{Sparse repair} means removing harmful judges on validation data and retraining the same aggregator on the selected subset. In short, \methodshort reads each judge's full distribution, decides how much to trust each judge for the current example, pools the distributions, and then computes one expected score.

\subsection{Judge Representations}

Each judge produces a normalized probability vector over the allowed score labels. Hard scores decode the argmax, soft scores take the expectation, and \methodshort uses the distribution itself. We optionally augment this bucket vector with scalar summaries, including hard and soft scores, confidence, entropy, and maximum bucket probability. Scores and targets are normalized to $[0,1]$ for cross-scale comparison.

This is the main distinction from score-level aggregation. Scalar baselines aggregate decoded scores or expected scores, so the aggregator only sees compressed judge outputs. \methodshort keeps the bucket vector as the input object and lets the gate decide whether distributional differences matter for the current example. Feature ablations in the appendix show that much of the useful signal lies in the score distribution itself, with scalar uncertainty features adding a smaller gain over probabilities alone.

\subsection{Aggregation Families}

For deployment, we use validation labels to compare a small family of gating and pooling variants. The sweep includes feed-forward and contextual gates inspired by mixture-of-experts gating~\citep{jacobs1991adaptive,jordan1994hierarchical}, Deep Sets-style permutation-invariant gating~\citep{zaheer2017deepsets}, and probability fusion with log-opinion pooling~\citep{heskes1997selecting}. We use this reproducible family to test whether aggregating judge score distributions before score compression improves evaluation quality.

The full validation comparison is shown in Appendix Table~\ref{tab:appendix-architecture-selection}. The selected mainline uses a DeepSets gate with tempered log-opinion pooling ($T=0.7$). For each example $x$, the gate assigns an example-specific weight to each judge from its distributional features. The following equations define the selected DeepSets gate and tempered log-opinion pool.

Let $\phi^{(j)}(x)\in\mathbb{R}^{d_\phi}$ be the feature vector of judge $j$, and let the shared encoder $\psi:\mathbb{R}^{d_\phi}\rightarrow\mathbb{R}^{d_h}$ map each judge feature vector into a common hidden space, $h^{(j)}(x)=\psi(\phi^{(j)}(x))$. With a shared scoring vector $a\in\mathbb{R}^{d_h}$, the judge weights are
\begin{equation}
\alpha^{(j)}(x)=\mathrm{softmax}_j\!\left(a^\top h^{(j)}(x)\right).
\end{equation}

We use log-opinion pooling because multiplicative fusion rewards agreement across judges. The trade-off is increased sensitivity to any judge that assigns near-zero mass to the correct bucket: a single near-zero probability can push a bucket down unless that judge receives very small weight. The weights parameterize the tempered log-opinion pool:
\begin{equation}
\begin{aligned}
q_k(x) &\propto \exp\!\Bigl(
\frac{1}{T}\sum_{j=1}^{m} \alpha^{(j)}(x)\cdot \log\!\left(p^{(j)}_k(x)+\varepsilon\right)
\Bigr),
\end{aligned}
\end{equation}
where $q_k(x)$ is the fused probability of score $s_k$, $T$ is the pooling temperature, and $\varepsilon$ is a small numerical constant added inside each judge log-probability for numerical stability. Lower temperatures sharpen the fused distribution toward buckets supported by the weighted pool, while higher temperatures flatten it. The final prediction is the expected score under the pooled distribution,
\begin{equation}
\hat{y}(x) = \sum_{k=1}^{K} s_k q_k(x).
\end{equation}
Compared with linear pooling, this operator sharpens consensus and is often stronger. Its brittleness motivates the sparse-repair analysis in Section~\ref{sec:operating-conditions}. In practice, the validation sweep over $T \in \{0.7, 1.0, 1.3\}$ selected $T=0.7$. We use this configuration because it gives the best validation average while maintaining a small parameter count and judge-order invariance.

\subsection{Training Objectives}

\paragraph{Direct and auxiliary objectives.}
The primary evaluator is selected on direct-score quality; pairwise supervision is used only as an auxiliary analysis. We train primarily on direct-score labels with Huber loss:
\begin{equation}
\mathcal{L}_{\text{direct}} = \mathrm{Huber}\bigl(\hat{y}(x), y\bigr).
\end{equation}

In auxiliary experiments, we add a margin-ranking loss on pairwise preference data. For a pair $(x_a, x_b)$ with human preference label $t \in \{-1,+1\}$ and scalar aggregator scores $s_a,s_b$, this loss is
\begin{equation}
\begin{aligned}
\mathcal{L}_{\text{pair}} =
\max\!\left(0, m - t\,(s_a - s_b)\right),
\end{aligned}
\end{equation}
where $m$ is a margin hyperparameter. The joint objective is
\begin{equation}
\mathcal{L} = \mathcal{L}_{\text{direct}} + \lambda \mathcal{L}_{\text{pair}}.
\end{equation}

\section{Experimental Setup}

Our experiments consist of two parts. First, the protocol study examines how scoring protocols change judge output distributions. Second, the aggregation benchmark fixes the candidate judges and compares aggregation methods on direct-score cells. The original aggregation benchmark contains 10 dataset--scale cells. After the judge pool, architecture, feature construction, and training procedure had been fixed, we evaluated the same configuration on six additional cells from HelpSteer2 validation, FLASK, and WildBench. We report the original 10 cells and the six additional cells separately, and use all 16 cells for the expanded comparison.

\begin{table}[t]
\centering
\footnotesize
\setlength{\tabcolsep}{2pt}
\hspace{-.38cm}
\resizebox{.47\textwidth}{!}{
\begin{tabular}{@{}p{2.2cm}p{2.5cm}r@{}}
\toprule
Dataset & Task & Size \\
\midrule
SummEval & Summarization & 6{,}400 \\
TopicalChat & Dialogue quality & 2{,}160 \\
ROSCOE-eSNLI & Reasoning quality & 151 \\
HelpSteer2 & Helpfulness scoring & 101{,}620 main \\
WMT MQM & Translation quality & 150{,}347 \\
\midrule
MT-Bench-human & Pairwise pref. & 2{,}575 eval \\
SFT validation & Pairwise pref. & 5{,}000 eval / 20{,}000 train \\
\bottomrule
\end{tabular}}
\caption{Datasets used in the paper. Direct-score datasets are evaluated with judge-score correlations, while pairwise datasets are used for auxiliary supervision and preference evaluation. The MT-Bench-human count is the filtered non-tie evaluation set; the SFT row lists the validation set used for pairwise evaluation and the training subset used for the auxiliary pairwise loss. Split-level counts appear in Appendix Table~\ref{tab:appendix-data-provenance}.}
\label{tab:dataset-summary}
\end{table}

\subsection{Tasks and Datasets}

Separate follow-ups then test heterogeneous experts, pairwise auxiliary training, and sparse repair.
We evaluate \methodshort on direct-score and pairwise benchmarks spanning summarization (SummEval~\citep{fabbri-etal-2021-summeval}), dialogue/helpfulness (TopicalChat-USR~\citep{gopalakrishnan2019topical,mehri-eskenazi-2020-usr} and HelpSteer2~\citep{wang2024helpsteer2}), reasoning/explanation quality (ROSCOE-eSNLI~\citep{camburu2018esnli,golovneva2023roscoe}), translation quality estimation (WMT MQM~\citep{freitag-etal-2021-experts,burchardt-2013-multidimensional}), and pairwise preference evaluation (MT-Bench-human~\citep{zheng2023judging} and Summarize-from-Feedback validation sets~\citep{stiennon2020learning}). These datasets include both high-resource cells where aggregation can be estimated reliably and smaller diagnostic cells where failure modes are visible. For HelpSteer2, Table~\ref{tab:dataset-summary} reports the main subset used in the original 10-cell benchmark; the held-out validation extension uses a separate source validation split summarized in the appendix.

\paragraph{Split integrity.}
To prevent leakage across train, validation, and test splits, we preserve the original split labels for each dataset and use exact example-ID intersections only when constructing aligned subset comparisons. Model, judge-pool, and aggregation choices are made using validation labels only; test labels are used only for final reporting.

\subsection{Judge Pool}

Our candidate pool includes six small judges from multiple model families: Qwen2.5-7B-Instruct~\citep{yang2024qwen25}, Qwen3.5-9B~\citep{qwen2026qwen35}, Gemma-2-9B-IT~\citep{gemmateam2024gemma2}, Gemma-3-12B-IT~\citep{gemmateam2025gemma3}, Llama-3.1-8B-Instruct~\citep{grattafiori2024llama3}, and Ministral-3-8B-Instruct-2512~\citep{liu2026ministral3}. The mainline comparison uses all six judges. The compact 3-judge configuration retains Gemma-2-9B-IT, Gemma-3-12B-IT, and Qwen2.5-7B-Instruct; this pool is selected once from the protocol study and then held fixed for the aggregation benchmark. A family-diverse 3-judge control enforces three different model families across the selected judges; it is weaker and appears in the appendix. The pool is homogeneous in evaluator type because all members are local instruct-tuned LLM judges; model families vary. Heterogeneous follow-ups append one task-aligned expert at a time to stress-test whether validation keeps, rejects, or isolates a candidate expert.

\subsection{Score Ranges and Prompting}

We evaluate score ranges [1--3], [1--5], [1--7], [1--10], and [1--100] as a diagnostic study of protocol sensitivity. The protocol study checks whether scoring choices change the measurement; the aggregation experiments then freeze a shared protocol to isolate pooling from repeated scale search. The main \methodshort protocol uses a shared scale across judges, with 1--5 as the default and 1--10 as the primary ablation. These scales serve as fixed deployment settings for aggregation comparisons. We also run a held-out protocol study over 24 dataset--model cells in which the protocol is selected on validation and evaluated once on test.

For each allowed score label, we extract or score the corresponding label probability, normalize across the allowed labels, and use the expectation as the soft score; hard decoding takes the highest-probability label. Soft-vs-hard counts exclude runs where rank correlation is undefined, such as constant prediction or target vectors; the appendix gives the exact filtering rule.

Prompts are task-specific but structurally aligned so the comparison stays focused on scale and aggregation. The appendix summarizes prompt skeletons and the shared output contract; the released artifact contains the exact prompt strings and parsing scripts.



\subsection{Baselines}

Baselines include the best single judge, uniform average or log-opinion pooling, covariance-aware aggregation, aligned strong single-judge follow-ups on selected cells, and \methodshort variants. Unless stated otherwise, ``best single'' denotes the strongest single judge within the evaluated cell; validation-selected single-judge results are reported with the deployment-selection diagnostics. The covariance-aware baseline is the strongest fixed pool: it uses validation error covariance to set non-uniform judge weights, giving less influence to judges with large or highly correlated errors, while keeping those weights fixed across examples.

\subsection{Implementation Details}

All aggregation experiments are script-driven and use fixed split manifests, cached judge score distributions, and reproducible checkpoint configurations. Appendix Tables~\ref{tab:appendix-reproducibility-packet} and~\ref{tab:appendix-implementation} summarize the hyperparameters, seed settings, prompt templates, runner scripts, and result files needed to reproduce the experiments.



\subsection{Metrics}

Before aggregation and evaluation, direct-score outputs are normalized to a common $[0,1]$ range to make scores comparable across datasets with different scales. For direct-score cells from SummEval, TopicalChat-USR, HelpSteer2, ROSCOE-eSNLI, and WMT MQM, we use Spearman correlation with human labels as the primary metric. Pairwise preference cells are evaluated with pairwise accuracy. We report MAE and RMSE only when absolute score error is directly relevant, such as in calibration or error-magnitude analyses. Main-paper tables report equal-cell averages, where a cell is one dataset paired with one score scale and is treated as one deployment condition. Sample-weighted summaries use the number of aligned evaluation examples in each cell as weights. For comparisons between aggregation methods, we report the mean cell-level difference and apply a Wilcoxon signed-rank test to the cell-level differences. One-sided tests evaluate whether \methodshort improves over the baseline, and two-sided tests evaluate whether the differences are nonzero. Some cells share datasets and examples across score scales, so these tests summarize variation across evaluation conditions with dependence between related cells.
\section{Main Results}
\label{sec:main-results}

We first show that scoring protocols change judge measurements, then test example-dependent weighting against fixed and single-judge baselines. The failures identify when validation should prefer fixed pooling, sparse repair, or a standalone expert.

\subsection{Protocol Choices Change the Measurement}

\begin{figure*}[t]
\centering
\includegraphics[width=2\columnwidth]{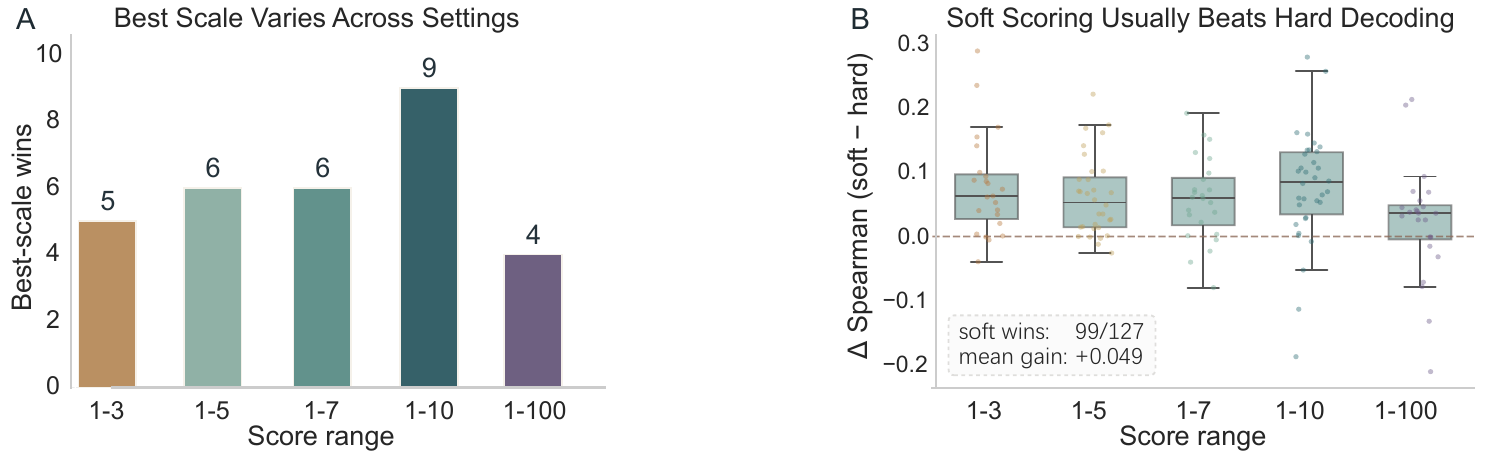}
\caption{Protocol sensitivity. Left: the best score range varies across dataset--model cells. Right: soft scoring usually improves over hard decoding across valid runs and scales.}
\label{fig:phase1-instability}
\end{figure*}

The protocol study shows that score range and extraction rule change judge behavior. Figure~\ref{fig:phase1-instability} shows two patterns: the best score range varies across dataset--model combinations, and soft scoring improves over hard decoding in most valid runs. These results support aggregating score distributions rather than scalar summaries alone. Across 24 dataset--model cells, validation-selected protocols also outperform fixed 1--5 or 1--10 soft scoring on test (Appendix Table~\ref{tab:appendix-protocol-study}).

\begin{table}[t]
\centering
\scriptsize
\setlength{\tabcolsep}{1pt}
\begin{tabular}{@{}lcccccc@{}}
\toprule
Comparison & Baseline & \methodshort & $\Delta$ & 95\% CI & \shortstack{one-sided\\$p$} & \shortstack{two-sided\\$p$} \\
\midrule
Uniform log & 0.304 & \best{0.383} & \best{+0.079} & [0.039, 0.119] & 0.005 & 0.010 \\
Cov-log & 0.341 & \best{0.383} & +0.043 & [-0.001, 0.085] & 0.053 & 0.105 \\
Best single & 0.341 & \best{0.383} & +0.042 & [-0.007, 0.088] & 0.053 & 0.105 \\
\bottomrule
\end{tabular}
\caption{Main 10-cell baseline comparisons for the selected 6-judge \methodshort checkpoint. \methodshort shows its largest and statistically significant gain over uniform log pooling. Comparisons to covariance-aware fixed pooling and the best single-judge baseline have positive mean gaps, but their confidence intervals cross zero and are not statistically significant. Confidence intervals are paired bootstrap intervals over cells. The last two columns report one-sided and two-sided paired Wilcoxon signed-rank test p-values over cells, without multiplicity correction.  ``Best single'' is the strongest single judge within each evaluated cell.}
\label{tab:phase2-main}
\end{table}

\subsection{Conditional Aggregation}

Table~\ref{tab:phase2-main} compares the selected configuration with equal-weight distributional pooling, fixed scalar baselines, and the best individual judges on the original 10-cell benchmark.

\paragraph{Gain over uniform pooling.}
The most direct comparison is uniform log pooling, which uses the same six judges with equal weights. \methodshort gains +0.079 mean Spearman, with Wilcoxon signed-rank tests giving $p=0.005$ for the one-sided improvement test and $p=0.010$ for the two-sided difference test. This supports example-specific weighting over equal-weight distributional pooling for this judge pool.

\paragraph{Comparison with stronger baselines.}
Against covariance-aware log pooling and the best single judge, \methodshort has positive mean gains, but the confidence intervals cross zero (Table~\ref{tab:phase2-main}). We therefore treat these comparisons as inconclusive and use validation data to choose among learned weighting, fixed pooling, and single-judge deployment.

\paragraph{Expanded evaluation.} We applied the same JudgeMoE configuration to six additional cells from HelpSteer2 validation, FLASK, and WildBench after the judge pool, architecture, features, and training procedure had been fixed. Across all 16 cells, JudgeMoE improves over the strongest single judge in the local pool by $+0.0393$ mean Spearman, with positive differences in 12/16 cells and a one-sided Wilcoxon signed-rank $p=0.0091$ ($p=0.0182$ two-sided). The mean difference from uniform log pooling is $+0.0601$ with 12/16 positive cells ($p=0.0017$ one-sided), while the difference from covariance-aware log pooling is $+0.0250$ with 9/16 positive cells ($p=0.0964$ one-sided). The six additional cells are reported individually in Appendix Tables~\ref{tab:appendix-helpsteer2val-mainline-extension}, \ref{tab:appendix-flask-mainline-extension}, and~\ref{tab:appendix-wildbench-direct}.

\paragraph{Cell-level variation.}
Appendix Table~\ref{tab:appendix-mainline-cellwise-significance} shows that \methodshort improves over the best single judge in most cells, but fails on both TopicalChat score ranges and WMT MQM@1--5. Section~\ref{sec:operating-conditions} uses these failures diagnostically: TopicalChat favors fixed pooling, while WMT MQM@1--5 motivates sparse judge selection and retraining.

\paragraph{Supporting analyses.} 
The appendix gives three additional analyses. First, excluding ROSCOE, the 8-cell average remains higher for \methodshort than for the best single judge (0.418 vs.\ 0.384). On the original 10 cells, the sample-weighted mean differences are $+0.0413$ relative to the best single judge and $+0.0549$ relative to covariance-aware log pooling; excluding ROSCOE gives $+0.0413$ and $+0.0548$, respectively (Appendix Table~\ref{tab:appendix-weighted-robustness}). Second, ablations show that the compact 3-judge setting remains competitive and that full score distributions carry most of the useful signal (Appendix Tables~\ref{tab:appendix-architecture-selection} and~\ref{tab:appendix-feature-ablation}). In a separate $2\times2$ comparison using the same cells, judge manifests, feature construction, pooling temperature, 30-epoch schedule, and three random seeds, DeepSets with log pooling achieves the highest mean Spearman of 0.4068 (Appendix Table~\ref{tab:gate_pooling_ablation}). Third, \methodshort obtains lower equal-cell ECE for the final scalar predictions than single-judge and fixed-pooling baselines (0.059 vs.\ 0.179--0.221; Appendix Table~\ref{tab:appendix-calibration}). This ECE compares normalized scalar predictions with human scores and does not measure calibration of the underlying score-token probabilities.

\subsection{Focused Heterogeneous Stress Tests}

For each heterogeneous follow-up, we append one task-aligned expert to the compact homogeneous pool. Validation labels choose among the homogeneous pool, the heterogeneous pool, a repaired subset, and the standalone expert when applicable. Table~\ref{tab:heterogeneous-followups} reports test Spearman correlations and validation-selected deployment choices.

\begin{table}[t]
\centering
\scriptsize
\setlength{\tabcolsep}{1.5pt}
\begin{tabular}{@{}lccc p{1.65cm}@{}}
\toprule
Follow-up & \shortstack{Hom.\\JudgeMoE} & \shortstack{Het.\\JudgeMoE} & Best single & \shortstack{Validation\\decision} \\
\midrule
SummEval + BERTSc. & 0.646 & \best{0.670} & 0.517 & keep expert \\
HelpSteer2 + RM & 0.533 & \best{0.545} & 0.390 & repair pool \\
WMT + trans. BERTSc. & \best{0.239} & 0.133 & 0.190 & keep LLM pool \\
WMT + COMET & 0.247 & 0.162 & \best{0.4326} & use standalone \\
\bottomrule
\end{tabular}
\caption{Focused heterogeneous follow-ups on compact deployment settings. All entries are test Spearman correlations with human labels; higher is better. BERTSc. denotes BERTScore and RM denotes the reward model. The validation decision is chosen using validation labels before test reporting. ``Keep expert'' selects the heterogeneous pool, ``repair pool'' selects a validation-repaired subset, ``keep LLM pool'' rejects the added expert, and ``use standalone'' keeps the external metric outside \methodshort.}
\label{tab:heterogeneous-followups}
\end{table}

\textbf{Positive additions.}
BERTScore helps on SummEval@1--10, and a Skywork reward model helps on held-out HelpSteer2 after train-split subset repair. These cases show that task-aligned experts can help when validation confirms complementary signal.

\textbf{Rejected additions.}
Translation metrics give the stronger warning. On WMT MQM@1--10, translation-BERTScore is rejected by subset search. COMET is the strongest standalone evaluator at 0.4326, while mixed-pool \methodshort reaches 0.1620. Validation selects COMET as the standalone evaluator. Strong standalone performance therefore does not necessarily translate into strong performance after adding the evaluator to the learned pool.

\section{Validation-Guided Deployment}
\label{sec:operating-conditions}

\subsection{Deployment Decision Matrix}

The previous section reports the evidence used for deployment decisions: protocol choice changes judge behavior, learned weights improve over uniform pooling, TopicalChat favors fixed pooling, and WMT MQM@1--5 improves after pruning and retraining. Table~\ref{tab:deployment-summary-main} summarizes how these validation patterns guide the deployment choice for a new judge pool.

\begin{table}[t]
\centering
\small
\setlength{\tabcolsep}{8pt}
\begin{tabular}{@{}ll@{}}
\toprule
Observed pattern & Choice \\
\midrule
Protocol variation & Select protocol \\
Strong fixed pool & Fixed pooling \\
Learned-weight gain & Full \methodshort \\
Weak judge & Subset + retraining \\
Strong external evaluator & Standalone evaluator \\
\bottomrule
\end{tabular}
\caption{Validation-guided deployment choices supported by the protocol study, main aggregation experiments, TopicalChat and WMT analyses, and heterogeneous-expert experiments.}
\label{tab:deployment-summary-main}
\end{table}

\textbf{Shared-error check.} Judges often make similar mistakes, which leaves less room for ensembling to help. Across the 10 direct-score cells, the mean judge-error correlation is 0.7775 for the full 6-judge pool and 0.7071 for the strong 3-judge pool (Appendix Table~\ref{tab:residual-correlation-bridge}). Values around 0.7 to 0.8 indicate that much of the remaining error is shared across judges.

\textbf{Validation selector.} The shared-error check motivates a validation-based choice among full, fixed, sparse, and single-judge modes. A thresholded selector matches the test-time best aggregation family in 6/8 held-out direct-score cells (Appendix Table~\ref{tab:deployment-rule-validation}). A stronger validation-only selector reaches 0.3732 average test Spearman across the eight held-out direct-score cells, close to the 0.3789 obtained by selecting the best family using test performance. Its selected fixed subsets use 3.38 judges on average (Appendix Table~\ref{tab:subset-selector-validation}).

 Calibration checks also favor \methodshort over the strongest single and fixed-pool baselines on dataset averages.

The focused heterogeneous follow-ups support the same conclusion. Task-aligned experts can help when they add new evidence to the homogeneous pool and pass train-split diagnostics, while prompt-diverse expansions of the same base models hurt, and both translation-BERTScore and COMET degrade the WMT mixed pool. This supports validating candidate experts before they enter the pool.

\subsection{Cell-Level Patterns}

The cell-wise breakdown explains why deployment should be selected on validation data. Most cells favor \methodshort, but TopicalChat and WMT MQM@1--5 show different failure patterns. TopicalChat is a fixed-pooling case: covariance-aware log pooling is already strong and residual judge-error correlation is extremely high. WMT MQM@1--5 is different because a weak judge contaminates the pool, so removing that judge and retraining the aggregator improves the result. These two cells motivate the fixed-pooling and sparse-repair rows in Table~\ref{tab:deployment-summary-main}.

\begin{figure}[t]
\centering
\includegraphics[width=0.76\columnwidth]{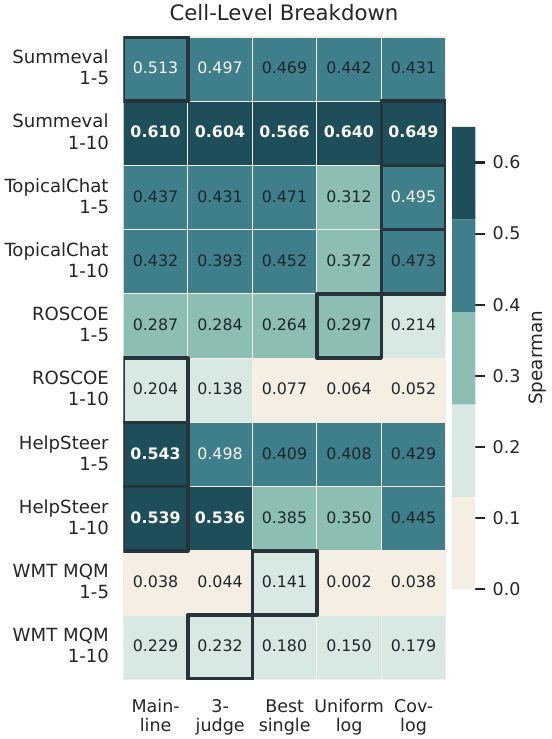}
\caption{Per-cell pattern for the main direct-score benchmark. \methodshort improves over the best single judge in most direct-score cells, but the per-cell breakdown also identifies failures on TopicalChat and WMT MQM@1--5 (Appendix Table~\ref{tab:appendix-mainline-cellwise-significance}). TopicalChat is a case where covariance-aware fixed pooling already outperforms \methodshort, while WMT MQM@1--5 motivates sparse repair: retraining \methodshort on the validation-selected sparse subset improves Spearman from 0.0376 to 0.1648 (Tables~\ref{tab:phase2-dataset-breakdown} and~\ref{tab:wmt-failure}). }
\label{fig:cell-patterns}
\end{figure}

\subsection{WMT Failure and Sparse Repair}

The WMT MQM diagnostic isolates a concrete failure mode: coarse score ranges compress ranking information, weak judges can hurt log pooling, and sparse task-specific composition can outperform all-judge fusion. At WMT 1--5, the full-6 log pool degrades sharply and the learned mainline only partially recovers (Table~\ref{tab:phase2-dataset-breakdown}). Leave-one-out validation identifies Qwen2.5-7B as a harmful weak judge. Keeping only the validation-selected top-2 subset and retraining sparse \methodshort raises WMT 1--5 to 0.1648, above both the best single judge at 0.1413 and the fixed sparse pool at 0.1473 (Table~\ref{tab:wmt-failure}). The full gate already assigns negligible weight to Qwen2.5 (Appendix Table~\ref{tab:appendix-wmt-gate-behavior}). Validation and test subset ranks are tightly aligned across all 120 WMT fixed-pool candidates, which consist of all single judges plus all non-single subsets under uniform-log and cov-log pooling (Appendix Table~\ref{tab:appendix-wmt-subset-stability}).

The heterogeneous translation follow-ups show the same failure mode. Appending translation-BERTScore drops the strong 3-judge \methodshort baseline from 0.2389 to 0.1328, and train-split subset search ranks the pure [Gemma-2, Gemma-3] subset first. COMET is more instructive: it becomes the strongest single expert at 0.4326 and uniform log pooling rises to 0.3926, while mixed-pool \methodshort still falls to 0.1620. Adding a strong external metric can therefore degrade the learned aggregator when its distributional behavior mismatches the LLM judges it was trained with. The conservative WMT recommendation is either a small clean LLM pool or the standalone metric itself. The appendix reports the full failure-and-recovery numbers.

The practical value comes from the reusable evidence layer as well as the full six-judge model. Once frozen judge distributions are cached, validation can choose among single, fixed, sparse, and gated modes without rerunning a monolithic evaluator.

\subsection{Implications for Evaluation Practice}

The results suggest a shift in how LLM judges are reported and reused. Individual judge quality remains important, yet the scoring protocol is part of the measurement and belongs with the judge output. A minimal benchmark report includes at least the score range, the score-extraction rule, and whether expected scores were computed from token probabilities or from decoded labels.

\paragraph{Probability access.} We also compare variants using different levels of judge output information. Hard-score-only features obtain 0.2634 mean Spearman, soft-score-only features obtain 0.3100, hard and soft scores obtain 0.3671, and probability-derived scalar summaries obtain 0.3776. The difference between hard and soft scores together and hard scores alone is $+0.1036$ with a one-sided Wilcoxon signed-rank $p=0.0068$, while the additional gain from probability-derived summaries over hard and soft scores is $+0.0106$ with $p=0.0801$. Full results appear in Appendix Table~\ref{tab:access_ablation}.
\section{Conclusion}

LLM-as-a-judge evaluation is best understood as protocol-conditioned measurement. The final score is only a compressed view of a distribution shaped by the scoring protocol. Our results show that score ranges and extraction rules change judge behavior and that example-specific distributional aggregation improves over equal-weight log pooling on the original 10-cell benchmark. Across the expanded 16-cell evaluation, \methodshort improves over the strongest local single judge by $+0.0393$ mean Spearman, with positive differences in 12/16 cells and a one-sided Wilcoxon signed-rank $p=0.0091$. The WMT-COMET case further shows that validation can favor a strong standalone evaluator when adding it to the learned pool reduces performance. The practical question is which cached measurements to trust, pool, prune, or keep standalone.
\section*{Limitations}

The original aggregation benchmark contains 10 dataset--scale cells, followed by six additional cells evaluated with the same \methodshort configuration. The resulting 16-cell analysis covers more evaluation conditions, while several cells still share datasets, score scales, and judge families. Evaluation on additional datasets and criteria would further establish how the observed gains transfer across applications.

This paper studies learned aggregation over a fixed pool of small local judges, so results depend on candidate quality, task match, and score resolution. Larger evaluators are included as standalone comparison points, while aggregation over pools of larger or proprietary judges remains for future study. Broader comparisons across code execution, factuality checking, and other evaluation settings would also require careful cost normalization \citep{huang2025impeach}.

The feature design is intentionally conservative. Current ablations suggest that judge distributions carry much of the useful signal. Comparisons with hidden states, explicit cross-judge interaction features, full natural-language explanations, and a broader same-capacity scalar-gated control remain useful next steps for separating representation gains from model-capacity gains.

The distributional approach also depends on access to informative judge score probabilities. The restricted-access experiments show that expected scores recover substantial signal beyond decoded scores, while probability-derived summaries provide a smaller additional improvement. The reported ECE measures the calibration of the final scalar predictions against human scores and does not establish calibration of the underlying score-token probabilities. In addition, the present benchmarks provide scalar human targets; annotator disagreement and multimodal human preferences remain outside the current scope.

\section*{Ethics Statement}

Automatic evaluation systems can influence model selection, training feedback, and deployment decisions, so errors or biases in judge models may propagate directly into research conclusions. Judge aggregation adds its own risks: protocol choices can privilege some score distributions over others, and weak-judge contamination can distort pooled rankings. We therefore emphasize comparison against human annotations, cross-dataset evaluation, explicit discussion of failure modes, and clear separation between architecture selection, held-out checkpoint expansion checks, and validation-only deployment heuristics. The appendix summarizes benchmark provenance and split counts; the released artifact records original license constraints and external API use where applicable. \methodshort is best understood as an efficiency-oriented evaluation aid, with careful human assessment still required in high-stakes settings.

\section*{Acknowledgements}
Joseph James was supported by the UKRI AI Centre for Doctoral Training in Speech and Language Technologies (SLT) and their Applications funded by UK Research and Innovation [grant number EP/S023062/1].

\bibliography{custom,anthology}

\clearpage
\appendix
\section{Background and Problem Setup}

We consider an evaluation instance $x$ consisting of task-specific context and a candidate output, with human target signal $y \in [0,1]$. Judge $j$ produces a categorical distribution $p^{(j)}(\cdot \mid x)$ over a finite score range $\mathcal{S} = \{s_1, \dots, s_K\}$, where the score values are normalized to $[0,1]$ before aggregation and evaluation.

\paragraph{Hard score.}
The hard score is the decoded label:
\begin{equation}
\hat{y}^{(j)}_{\mathrm{hard}} = \arg\max_{s \in \mathcal{S}} p_j(s \mid x).
\end{equation}

\paragraph{Soft score.}
The soft score uses the expected score under the judge distribution:
\begin{equation}
\hat{y}^{(j)}_{\mathrm{soft}} = \sum_{s \in \mathcal{S}} s \, p_j(s \mid x).
\end{equation}

Our experiments compare both protocols across several score ranges. The empirical motivation for \methodshort is that the optimal score range is unstable across judges and tasks, while each judge carries complementary information through its full score distribution, confidence, and entropy. This motivates an aggregation layer that consumes score distributions directly while preserving information that would be discarded after decoding a single label.

\paragraph{Goal.}
Given multiple judges $\{j_1, \dots, j_m\}$, we want to learn an aggregation function
\begin{equation}
\hat{y}(x) =
f\bigl(
z^{(1)}(x), \dots, z^{(m)}(x)
\bigr),
\end{equation}
where each $z^{(j)}(x)$ includes score-distribution features and optional metadata. The desired aggregator is lightweight, easy to train, and compatible with both direct-score and pairwise supervision.

\section{Additional Experimental Settings and Results}

\paragraph{Roadmap.}
The appendix first gives prompt templates and artifact accounting, then reports the full per-cell and deployment-rule tables behind the main results. It next provides implementation and selection details, pairwise and out-of-domain checks, qualitative and calibration diagnostics, compute proxies, and finally the theoretical propositions used to interpret the operating conditions.

\subsection{Prompt Templates}

Table~\ref{tab:appendix-prompt-schema} summarizes the prompt policy used throughout the project, and Table~\ref{tab:appendix-prompt-templates} gives the default direct-score template skeletons. The exact source and candidate fields differ across summarization, dialogue, reasoning, and translation, but the output contract is deliberately uniform so that scale and extraction effects can be isolated cleanly.

\begin{table}[t]
\centering
\small
\setlength{\tabcolsep}{8pt}
\begin{tabular}{@{}ll@{}}
\toprule
Component & Prompt content \\
\midrule
Instruction & Evaluator role + criterion \\
Context & Source / prompt / dialogue \\
Candidate & Output to score \\
Reference & Optional task reference \\
Output & Scalar score in specified range \\
\bottomrule
\end{tabular}
\caption{Shared prompt components for direct-score judging. Dataset-specific wording varies, while all prompts use the same scalar-score output format.}
\label{tab:appendix-prompt-schema}
\end{table}

\begin{table*}[t]
\centering
\scriptsize
\setlength{\tabcolsep}{10pt}
\renewcommand{\arraystretch}{1.06}
\begin{tabularx}{\textwidth}{@{}>{\bfseries\raggedright\arraybackslash}p{0.12\textwidth}>{\ttfamily\raggedright\arraybackslash}X@{}}
\toprule
Task family & \normalfont\bfseries Default direct-score template skeleton \\
\midrule
Summarization & You are an expert summarization evaluator. Evaluate the [aspect] of the generated summary. [Aspect-specific criterion.] SOURCE DOCUMENT: \{source\} GENERATED SUMMARY: \{candidate\} Return only a score from \{low\} to \{high\}. \\
\addlinespace[3pt]
Dialogue & You are an expert dialogue response evaluator. Evaluate the [aspect] of the assistant response. CONTEXT: \{source\} ASSISTANT RESPONSE: \{candidate\} Return only a score from \{low\} to \{high\}. \\
\addlinespace[3pt]
Assistant response & You are an expert evaluator of assistant responses. Evaluate the [aspect] of the response. USER REQUEST: \{source\} ASSISTANT RESPONSE: \{candidate\} Return only a score from \{low\} to \{high\}. \\
\addlinespace[3pt]
Reasoning & You are an expert evaluator of reasoning chains. Evaluate the overall quality of the reasoning chain. Consider correctness, completeness, and coherence. PROBLEM OR CONTEXT: \{source\} REASONING CHAIN: \{candidate\} Return only a score from \{low\} to \{high\}. \\
\addlinespace[3pt]
Translation & You are an expert translation evaluator. Evaluate the overall translation quality. Consider adequacy and fluency. SOURCE SEGMENT: \{source\} TRANSLATION: \{candidate\} [optional HUMAN REFERENCE] Return only a score from \{low\} to \{high\}. \\
\addlinespace[3pt]
FLASK-style & You are an expert evaluator of assistant responses. Evaluate the candidate response using the provided criterion and scoring rubric. INSTRUCTION: \{source\} CANDIDATE RESPONSE: \{candidate\} CRITERION: \{criterion\} [reference/rubric blocks] Return only a score from \{low\} to \{high\}. \\
\bottomrule
\end{tabularx}
\caption{Default direct-score prompt templates. All task families use the same scalar-score output format; prompt-style variants are evaluated separately in Table~\ref{tab:appendix-prompt-style}.}
\label{tab:appendix-prompt-templates}
\end{table*}

\subsection{Additional Tables}

Table~\ref{tab:appendix-artifacts} lists the experiment artifacts used to construct the paper tables. We keep the full benchmark matrices in the released artifacts and include compact summaries in the paper due to space constraints.

\begin{table}[t]
\centering
\small
\setlength{\tabcolsep}{8pt}
\begin{tabular}{@{}ll@{}}
\toprule
Experiment & Paper content \\
\midrule
Protocol study & Scale + soft/hard scoring \\
Aggregation & Main results + ablations \\
Seed sweep & 6-judge / 3-judge variation \\
Pairwise study & Direct + pairwise training \\
Large judges & Quality + efficiency \\
WMT MQM & Pool failure + sparse retraining \\
\bottomrule
\end{tabular}
\caption{Experiment groups and their corresponding analyses in the paper.}
\label{tab:appendix-artifacts}
\end{table}

\begin{table*}[t]
\centering
\small
\setlength{\tabcolsep}{8pt}
\begin{tabular}{@{}lllll@{}}
\toprule
Family & Aggregation input & Full distribution & Example-specific weights & Interaction \\
\midrule
Single-judge methods & Score / rationale / checklist & No & No & No \\
Jury / debate & Votes / answers / rationales & Usually no & Sometimes & Yes \\
Latent-reliability models & Labels / preferences & No & Judge-level & No \\
Scalar weighting & Scalar scores & No & Yes & No \\
Fixed opinion pooling & Score distributions & Yes & No & No \\
\methodshort & Cached score distributions & Yes & Yes & No \\
\bottomrule
\end{tabular}
\caption{Comparison of related evaluator aggregation methods. \methodshort combines full score distributions with example-specific judge weighting.}
\label{tab:appendix-related-taxonomy}
\end{table*}

\begin{table*}[t]
\centering
\small
\setlength{\tabcolsep}{6pt}
\begin{tabular}{@{}lrrrrrl@{}}
\toprule
Dataset & Train & Dev & Valid & Test & Total & Note \\
\midrule
\multicolumn{7}{@{}l}{\textit{\textbf{Direct-score corpus}}} \\
All direct-score data & 212{,}953 & 15{,}912 & 5{,}190 & 31{,}813 & 265{,}868 & -- \\
\addlinespace[2pt]
HelpSteer2  & -- & -- & -- & -- & 106{,}810 & Train / validation preserved \\
SummEval    & -- & -- & -- & -- & 6{,}400   & Human scores retained \\
TopicalChat & -- & -- & -- & -- & 2{,}160   & Dialogue aspects retained \\
ROSCOE      & -- & -- & -- & -- & 151       & Diagnostic subset \\
WMT MQM     & -- & -- & -- & -- & 150{,}347 & MQM metadata retained \\
\addlinespace[3pt]
\multicolumn{7}{@{}l}{\textit{\textbf{Pairwise corpus}}} \\
All pairwise data & 724{,}238 & 6{,}878 & 104{,}522 & 48{,}916 & 884{,}554 & MT-Bench ties removed \\
\bottomrule
\end{tabular}
\caption{Dataset composition and split counts for the released evaluation data. Licenses and use restrictions follow the original datasets and are recorded in the release manifest.}
\label{tab:appendix-data-provenance}
\end{table*}

\begin{table*}[t]
\centering
\small
\setlength{\tabcolsep}{7pt}
\begin{tabular}{@{}lll@{}}
\toprule
Item & Setting & Reference \\
\midrule
Architecture & DeepSets, $d_h{=}128$, dropout 0.1, log pool, $T{=}0.7$, $\varepsilon{=}10^{-6}$ & Table~\ref{tab:appendix-architecture-selection} \\
Optimization & Adam, lr $8{\times}10^{-4}$, batch 128, 20 epochs & Released logs \\
Objective & Huber loss; auxiliary ranking loss with margin 0.0 & Section~3.3, Table~\ref{tab:pairwise-main} \\
Seeds & 13; sweep: 13, 17, 29 & Table~\ref{tab:appendix-seeds} \\
Features & $K$ score probabilities + scalar summaries & Table~\ref{tab:appendix-feature-ablation} \\
Splits & Original split labels; exact example-ID intersection & Table~\ref{tab:appendix-data-provenance} \\
Release & Runners, manifests, prompts, configs, results & Release package \\
\bottomrule
\end{tabular}
\caption{Key settings and reproducibility references for the main aggregation experiments. The selected configuration is unchanged for the six additional evaluation cells.}
\label{tab:appendix-reproducibility-packet}
\end{table*}

\begin{table*}[t]
\centering
\footnotesize
\setlength{\tabcolsep}{4pt}
\begin{tabular}{@{}llcccccc@{}}
\toprule
Dataset & Scale & Mainline & 3-judge & Best single & Best judge & Uniform log & Cov-log \\
\midrule
SummEval & 1--5 & 0.5134 & 0.4967 & 0.4687 & Gemma-2-9B & 0.4420 & 0.4313 \\
SummEval & 1--10 & 0.6101 & 0.6043 & 0.5657 & Gemma-3-12B & 0.6401 & 0.6494 \\
TopicalChat & 1--5 & 0.4357 & 0.4297 & 0.4683 & Gemma-2-9B & 0.3108 & 0.4928 \\
TopicalChat & 1--10 & 0.4270 & 0.3925 & 0.4458 & Gemma-2-9B & 0.3700 & 0.4708 \\
ROSCOE & 1--5 & 0.2870 & 0.2841 & 0.2641 & Gemma-3-12B & 0.2975 & 0.2142 \\
ROSCOE & 1--10 & 0.2041 & 0.1381 & 0.0773 & Ministral-3-8B & 0.0644 & 0.0516 \\
HelpSteer2 & 1--5 & 0.5426 & 0.4977 & 0.4087 & Gemma-2-9B & 0.4076 & 0.4285 \\
HelpSteer2 & 1--10 & 0.5386 & 0.5356 & 0.3849 & Gemma-2-9B & 0.3497 & 0.4449 \\
WMT MQM & 1--5 & 0.0376 & 0.0443 & 0.1413 & Gemma-2-9B & 0.0016 & 0.0383 \\
WMT MQM & 1--10 & 0.2293 & 0.2316 & 0.1797 & Gemma-2-9B & 0.1504 & 0.1793 \\
\bottomrule
\end{tabular}
\caption{Per-dataset breakdown for the main aggregation systems. The gains vary across dataset and scale cells: the main gains come from SummEval at 1--5, HelpSteer, and WMT MQM at 1--10, while TopicalChat and SummEval at 1--10 remain challenging against strong fixed baselines. ROSCOE is reported as a small-sample diagnostic cell and is interpreted cautiously because it contains only 151 examples.}
\label{tab:phase2-dataset-breakdown}
\end{table*}

\begin{table}[t]
\centering
\small
\setlength{\tabcolsep}{4pt}
\begin{tabular}{@{}lccc@{}}
\toprule
Temperature & Avg. Sp. & $\Delta$ vs. single & Wins / 10 \\
\midrule
$T{=}1.0$ & 0.3828 & 0.0414 & 2 \\
$T{=}0.7$ & \best{0.3832} & \best{0.0419} & \best{4} \\
$T{=}1.3$ & 0.3729 & 0.0315 & 4 \\
\bottomrule
\end{tabular}
\caption{Temperature sensitivity for DeepSets + log pooling. The values are taken from the 3-way temperature comparison on the same 10 dataset--scale cells.}
\label{tab:temp-sensitivity}
\end{table}

\paragraph{Architecture variants.}
The validation sweep compares several gates for computing judge weights $w_j(x)$. The linear gate uses
\[
w(x)=\mathrm{softmax}(W\phi(x)+b),
\]
where $\phi(x)$ concatenates the per-judge features. The low-rank gate factorizes the weight matrix as $W=UV^\top$ to reduce parameters. The DeepSets gate computes permutation-invariant judge weights by applying a shared encoder to each judge feature vector and normalizing the resulting scores:
\[
h_j=\rho(\phi^{(j)}(x)), \qquad
w_j(x)=\mathrm{softmax}_j(a^\top h_j).
\]
For distributional variants, the final pooled distribution is computed with tempered log-opinion pooling,
\[
p_{\mathrm{pool}}(y \mid x) \propto 
\prod_{j=1}^{J} p_j(y \mid x)^{w_j(x)/T}.
\]

\begin{table}[t]
\centering
\footnotesize
\setlength{\tabcolsep}{4pt}
\begin{tabular}{@{}lc@{}}
\toprule
Learned aggregation configuration & Avg. Sp. \\
\midrule
Linear + log pool & 0.3511 \\
FFN + linear pool & 0.3505 \\
Low-rank linear + log & 0.3546 \\
DeepSets + log pool & 0.3828 \\
DeepSets + log, $T{=}0.7$ & \best{0.3832} \\
3-judge efficient & 0.3656 \\
\bottomrule
\end{tabular}
\caption{Learned-aggregation selection on the original 10-cell suite. The selected mainline is DeepSets gating with tempered log pooling at $T{=}0.7$; Table~\ref{tab:phase2-main} uses this fixed configuration for the main baseline comparisons.}
\label{tab:appendix-architecture-selection}
\end{table}

\begin{table}[t]
\centering
\small
\begin{tabular}{lcc}
\toprule
Gate & Linear pooling & Log pooling \\
\midrule
FFN & $0.3623 \pm 0.0016$ & $0.3751 \pm 0.0004$ \\
DeepSets & $0.3831 \pm 0.0026$ & $\mathbf{0.4068 \pm 0.0016}$ \\
\bottomrule
\end{tabular}
\caption{$2\times2$ gate and pooling comparison on the original 10 cells. All four settings use the same judge manifests, feature construction, pooling temperature, 30-epoch training schedule, and three random seeds. Values are mean Spearman $\pm$ standard deviation. This experiment uses a separate training schedule from the reported mainline checkpoint, so the absolute values are interpreted within this table.}
\label{tab:gate_pooling_ablation}
\end{table}

\begin{table}[t]
\centering
\footnotesize
\setlength{\tabcolsep}{4pt}
\begin{tabular}{@{}lcc@{}}
\toprule
Selector & Equal-cell Sp. & Weighted Sp. \\
\midrule
Validation-selected protocol & 0.1865 & 0.1674 \\
Fixed 1--5 soft & 0.1486 & 0.0269 \\
Fixed 1--10 soft & 0.1813 & 0.1002 \\
Best protocol by test performance & 0.2569 & 0.1681 \\
\bottomrule
\end{tabular}
\caption{Held-out protocol study over 24 dataset and model cells. The validation-selected protocol outperforms fixed 1--5 and 1--10 soft scoring on equal-cell Spearman. The selected protocols span all five tested score ranges, showing that protocol sensitivity is not limited to one fixed range.}
\label{tab:appendix-protocol-study}
\end{table}

\begin{table}[t]
\centering
\footnotesize
\setlength{\tabcolsep}{4pt}
\begin{tabular}{@{}lccc@{}}
\toprule
Style & Rows & Avg. soft Sp. & Wins vs. default \\
\midrule
Default & 12 & 0.3230 & 0/12 \\
Rubric & 12 & 0.2275 & 0/12 \\
Critical & 12 & 0.2968 & 4/12 \\
\bottomrule
\end{tabular}
\caption{Reduced prompt-style protocol check over two judges, three datasets, two scales, and three phrasing variants. A non-default prompt is best in 4/12 dataset--model--scale cells. The rubric wording is uniformly weaker than the default prompt, while the critical wording helps in a small subset of cases, including Qwen on WMT MQM (+0.1158 at 1--5 and +0.3023 at 1--10).}
\label{tab:appendix-prompt-style}
\end{table}

\begin{table}[t]
\centering
\footnotesize
\setlength{\tabcolsep}{3pt}
\begin{tabular}{@{}p{1.8cm}cccc@{}}
\toprule
Pool & Judges & Mean $\rho$ & JudgeMoE Sp. & Wins vs. default \\
\midrule
Default 2 & 2 & 0.7955 & 0.4049 & -- \\
Protocol-diverse 6 & 6 & 0.8472 & 0.3883 & 3 / 3 \\
\bottomrule
\end{tabular}
\caption{Protocol-diverse pool pilot on \texttt{summeval}, \texttt{topical\_chat}, and \texttt{wmt\_mqm} at \texttt{1--5} and \texttt{1--10}. The 6-judge pool augments a default 2-judge pool (\texttt{Qwen2.5-7B}, \texttt{gemma-2-9B}) with \texttt{rubric} and \texttt{critical} prompt variants of the same two models. Mean residual correlation increases, and average JudgeMoE performance drops slightly, indicating that added protocol diversity can fail to yield useful complementarity.}
\label{tab:protocol-diverse-pool}
\end{table}

\begin{table}[t]
\centering
\small
\setlength{\tabcolsep}{4pt}
\begin{tabular}{@{}lccc@{}}
\toprule
Feature set & Avg. Sp. & $\Delta$ vs. single & Wins / 10 \\
\midrule
Full current & \best{0.3832} & \best{0.0419} & \best{6} \\
Distribution + scores & 0.3811 & 0.0398 & 2 \\
Distribution only & 0.3746 & 0.0332 & 2 \\
\bottomrule
\end{tabular}
\caption{Feature ablation for the main DeepSets + log-pool configuration. Most of the signal already lies in the judge score distributions, but the full scalar summary bundle provides a small and consistent gain over using distributions alone.}
\label{tab:appendix-feature-ablation}
\end{table}

\begin{table}[t]
\centering
\small
\begin{tabular}{lcc}
\toprule
Input features & Mean Sp. & SD \\
\midrule
Hard score only & 0.2634 & 0.0028 \\
Soft score only & 0.3100 & 0.0088 \\
Hard and soft scores & 0.3671 & 0.0027 \\
Probability-derived scalar summaries & 0.3776 & 0.0053 \\
\bottomrule
\end{tabular}
\caption{Performance with different levels of access to judge outputs on the original 10 cells.}
\label{tab:access_ablation}
\end{table}

\begin{table}[t]
\centering
\footnotesize
\setlength{\tabcolsep}{3pt}
\begin{tabular}{@{}lcc@{}}
\toprule
Cell & Full-6 $\bar{\rho}$ & Strong-3 $\bar{\rho}$ \\
\midrule
Average (10 cells) & 0.7775 & 0.7071 \\
SummEval 1--10 & 0.7611 & 0.5860 \\
TopicalChat 1--5 & 0.8705 & 0.7875 \\
TopicalChat 1--10 & 0.9467 & 0.9150 \\
WMT MQM 1--10 & 0.5883 & 0.5696 \\
\bottomrule
\end{tabular}
\caption{Mean off-diagonal residual correlation $\rho$ among judge errors, computed from normalized soft scores relative to human labels. The large correlations on TopicalChat imply that most remaining error is shared across judges, which is consistent with Proposition~3 and helps explain why fixed covariance-aware pooling is already strong there.}
\label{tab:residual-correlation-bridge}
\end{table}

\begin{table}[t]
\centering
\footnotesize
\setlength{\tabcolsep}{4pt}
\begin{tabular}{@{}lccc@{}}
\toprule
Cell & Best test family & Selected & Correct \\
\midrule
ROSCOE 1--10 & sparse & sparse & yes \\
ROSCOE 1--5 & fixed & fixed & yes \\
SummEval 1--10 & fixed & fixed & yes \\
SummEval 1--5 & sparse & fixed & no \\
TopicalChat 1--10 & fixed & fixed & yes \\
TopicalChat 1--5 & fixed & fixed & yes \\
WMT MQM 1--10 & full & full & yes \\
WMT MQM 1--5 & sparse & full & no \\
\bottomrule
\end{tabular}
\caption{Leave-one-cell-out diagnostic for the deployment heuristic. We perform a coarse threshold sweep over $\rho$, leave-one-out gain, and the fixed-vs-full margin on the eight direct-score cells with held-out validation and test splits. All eight folds recover the same selector: choose sparse when the leave-one-out gain is at least 0.14; otherwise choose fixed when $\rho \ge 0.5$ and the fixed pool is already at least as strong as full JudgeMoE on validation; otherwise choose full. This rule matches the best test family on 6/8 cells.}
\label{tab:deployment-rule-validation}
\end{table}

\begin{table}[t]
\centering
\footnotesize
\setlength{\tabcolsep}{4pt}
\begin{tabular}{@{}lcc@{}}
\toprule
Strategy & Avg. test Sp. & Weighted test Sp. \\
\midrule
Best single (valid-selected) & 0.2748 & 0.1860 \\
Always fixed pool & 0.3374 & 0.1319 \\
Always full \methodshort & 0.2975 & 0.1655 \\
Always sparse variant & 0.3497 & 0.1536 \\
Best family by validation performance & 0.3402 & 0.1650 \\
LOO heuristic selector & 0.3495 & 0.1694 \\
Best family by test performance & 0.3639 & 0.1701 \\
\bottomrule
\end{tabular}
\caption{Validation-only strategy selection on the eight direct-score cells with held-out validation and test splits. Thresholds are selected on the remaining cells and then applied to the held-out test cell. The resulting selector improves over always using full \methodshort and over always using the strongest fixed pool on equal-cell average test Spearman, while sample-weighted averages remain dominated by the large WMT cells and therefore stay close to the best family by test performance.}
\label{tab:strategy-selector-validation}
\end{table}

\begin{table}[t]
\centering
\scriptsize
\setlength{\tabcolsep}{4pt}
\resizebox{\columnwidth}{!}{%
\begin{tabular}{@{}p{3cm}cc@{}}
\toprule
Strategy & Avg. test Sp. & Weighted test Sp. \\
\midrule
Best single (valid-selected) & 0.2748 & 0.1860 \\
Full fixed pool & 0.3374 & 0.1319 \\
Full \methodshort{} & 0.3154 & 0.1665 \\
Valid-selected subset fixed & 0.3133 & 0.2113 \\
Valid-selected family & 0.3732 & 0.2213 \\
Best family by test performance & 0.3789 & 0.2215 \\
\bottomrule
\end{tabular}
}
\caption{Validation-only subset selection on the eight held-out direct-score cells. The best fixed subset uses 3.38 judges on average and is smaller than the full six-judge pool in all eight cells. A validation-selected family over best single, full fixed pooling, full \methodshort{}, and subset-fixed pooling nearly matches the best family selected using test performance.}
\label{tab:subset-selector-validation}
\end{table}

\begin{table}[t]
\centering
\small
\setlength{\tabcolsep}{6pt}
\begin{tabular}{@{}p{0.62\columnwidth}p{0.28\columnwidth}@{}}
\toprule
Validation signal & Recommended choice \\
\midrule
High residual correlation; covariance-aware pooling already strong & Fixed covariance-aware pooling \\
\addlinespace
Large leave-one-out gain or large subset/full-pool gap & Subset selection + retraining \\
\addlinespace
Remaining gain from example-specific weighting & Full \methodshort \\
\bottomrule
\end{tabular}
\caption{Validation signals and corresponding deployment choices. Appendix Table~\ref{tab:deployment-rule-validation} reports the leave-one-cell-out evaluation.}
\label{tab:deployment-guidance}
\end{table}

\begin{table*}[t]
\centering
\small
\setlength{\tabcolsep}{6pt}
\begin{tabular}{@{}lll@{}}
\toprule
Stage & Validation signal & Selection \\
\midrule
Protocol & Scale / extraction differences & Select protocol \\
Family & Fixed / full / single performance & Select aggregation family \\
Subset & Subset performance & Select judge subset \\
Subset retraining & Subset / full-pool difference & Retrain on selected subset \\
\bottomrule
\end{tabular}
\caption{Validation procedure for protocol, aggregation-family, and judge-subset selection. Supporting results are reported in Appendix Tables~\ref{tab:appendix-protocol-study}, \ref{tab:deployment-rule-validation}, \ref{tab:strategy-selector-validation}, and~\ref{tab:subset-selector-validation}, and in Table~\ref{tab:wmt-failure}.}
\label{tab:deployment-playbook}
\end{table*}

\begin{table}[t]
\centering
\footnotesize
\setlength{\tabcolsep}{4pt}
\begin{tabular}{@{}lccc@{}}
\toprule
Subset & Weighted Sp. & $\Delta$ vs. single & $\Delta$ vs. cov-log \\
\midrule
All 10 cells & 0.3052 & 0.0413 & 0.0549 \\
Exclude ROSCOE & 0.3052 & 0.0413 & 0.0548 \\
Exclude WMT & 0.5393 & 0.1346 & 0.0961 \\
Non-HelpSteer & 0.1549 & -0.0241 & 0.0236 \\
\bottomrule
\end{tabular}
\caption{Sample-weighted summaries for the mainline 6-judge \methodshort system. On the full 10-cell suite, the weighted bootstrap 95\% interval is $[-0.0630, 0.1399]$ against the best single judge and $[0.0109, 0.1014]$ against covariance-aware pooling. Excluding ROSCOE leaves the weighted means essentially unchanged.}
\label{tab:appendix-weighted-robustness}
\end{table}

\begin{table*}[ht]
\centering
\scriptsize
\resizebox{.9\textwidth}{!}{
\begin{tabular}{@{}lrrrrrr@{}}
\toprule
Cell & $n$ & JudgeMoE & Best single & $\Delta$ & 95\% CI & $p$ \\
\midrule
helpsteer2\_train@1-10 & 101620 & 0.5386 & 0.3849 & 0.1537$^\dagger$ & [0.1502, 0.1572] & 0.0100 \\
helpsteer2\_train@1-5 & 101620 & 0.5426 & 0.4087 & 0.1339$^\dagger$ & [0.1303, 0.1373] & 0.0100 \\
roscoe\_esnli@1-10 & 151 & 0.2041 & 0.0773 & 0.1268 & [-0.0240, 0.2675] & 0.1294 \\
roscoe\_esnli@1-5 & 151 & 0.2870 & 0.2641 & 0.0228 & [-0.0063, 0.0465] & 0.1791 \\
summeval@1-10 & 6400 & 0.6101 & 0.5657 & 0.0444$^\dagger$ & [0.0260, 0.0653] & 0.0100 \\
summeval@1-5 & 6400 & 0.5134 & 0.4687 & 0.0448$^\dagger$ & [0.0319, 0.0591] & 0.0100 \\
topical\_chat@1-10 & 2160 & 0.4270 & 0.4458 & -0.0188 & [-0.0479,0.0095] & 0.1916 \\
topical\_chat@1-5 & 2160 & 0.4357 & 0.4683 & -0.0325 & [-0.0566,-0.0094] & 0.0074 \\
wmt\_mqm@1-10 & 150347 & 0.2293 & 0.1797 & 0.0495$^\dagger$ & [0.0454, 0.0524] & 0.0100 \\
wmt\_mqm@1-5 & 150347 & 0.0376 & 0.1413 & -0.1037 & [-0.1091, -0.0991] & 0.0100 \\
\midrule
\multicolumn{7}{@{}l@{}}{Positive $\Delta$ in 7/10 cells; significant positive $\Delta$ in 5/10 cells. $^\dagger p<0.05$ and CI excludes 0.} \\
\bottomrule
\end{tabular}}
\caption{Per-cell paired bootstrap significance for the main 10-cell aggregation comparison against the best single judge, using the exact per-example predictions of the reported mainline checkpoint. This supplements the aggregate 10-cell Wilcoxon test by showing which benchmark cells individually support the positive \methodshort trend.}
\label{tab:appendix-mainline-cellwise-significance}
\end{table*}

\begin{table}[t]
\centering
\footnotesize
\setlength{\tabcolsep}{5pt}
\begin{tabular}{@{}lcc@{}}
\toprule
System & ECE $\downarrow$ & MAE $\downarrow$ \\
\midrule
JudgeMoE & \best{0.0588} & \best{0.1890} \\
Best single & 0.2120 & 0.2782 \\
Uniform log & 0.1785 & 0.2718 \\
Cov-log & 0.2214 & 0.2976 \\
\bottomrule
\end{tabular}
\caption{Continuous-score calibration check on the original 10 direct-score cells. ECE uses 10 equal-width bins over normalized predictions; within each bin we compute the absolute gap between mean prediction and mean human score, then average these gaps by bin frequency. Values are equal-cell averages, matching the main Spearman aggregation.}
\label{tab:appendix-calibration}
\end{table}

\begin{table}[t]
\centering
\scriptsize
\setlength{\tabcolsep}{3pt}
\begin{tabular}{@{}lcccc@{}}
\toprule
Dataset & JudgeMoE & Best single & Uniform log & Cov-log \\
\midrule
SummEval & 0.0661 & 0.3840 & 0.1621 & 0.2210 \\
TopicalChat & 0.0559 & 0.2640 & 0.1385 & 0.2118 \\
ROSCOE & 0.1053 & 0.2161 & 0.2693 & 0.3381 \\
HelpSteer2 & 0.0133 & 0.0483 & 0.0859 & 0.0791 \\
WMT MQM & 0.0534 & 0.1476 & 0.2368 & 0.2568 \\
\bottomrule
\end{tabular}
\caption{Per-dataset ECE breakdown, averaged over the two main scales for each dataset. The aggregate calibration gain spans the full benchmark set: JudgeMoE has the lowest ECE on all five dataset averages in this diagnostic.}
\label{tab:appendix-calibration-by-dataset}
\end{table}

\begin{table}[t]
\centering
\footnotesize
\setlength{\tabcolsep}{4pt}
\begin{tabular}{@{}lcccc@{}}
\toprule
Comparison & Mean $\Delta$ & SD$_\Delta$ & Power @10 & Cells for 80\% \\
\midrule
vs. uniform log & +0.0794 & 0.0692 & 0.98 & 5 \\
vs. cov-log & +0.0427 & 0.0729 & 0.58 & 19 \\
vs. best single & +0.0419 & 0.0812 & 0.49 & 24 \\
\bottomrule
\end{tabular}
\caption{First-order power note for the original 10-cell paired comparisons. We use the observed cell-level mean and standard deviation of the paired deltas with a one-sided $\alpha{=}0.05$ normal approximation. The table explains why the 10-cell benchmark is sufficient for the uniform-log control but underpowered for the smaller gaps against covariance-aware pooling and the best single judge.}
\label{tab:appendix-power-note}
\end{table}

\begin{table*}[ht]
\centering
\small
\setlength{\tabcolsep}{7pt}
\begin{tabular}{@{}p{0.20\linewidth}p{0.42\linewidth}p{0.30\linewidth}@{}}
\toprule
Component & Selection setup & Reported use \\
\midrule
Protocol study & Five score ranges with hard/soft extraction; validation-based protocol selection over 24 cells & Select scoring protocol \\
Main architecture & Five architecture/pooling variants, temperature sweep, and feature ablation on the original 10 cells & Select 6-judge DeepSets + log pooling \\
Compact 3-judge pool & Candidate triples from protocol-study screening & Efficiency setting \\
WMT subset selection & Top-$k$ subsets and leave-one-out selection on WMT validation data & Retrain selected subsets \\
General subset selector & All 63 nonempty judge subsets over eight validation/test cells & Select judge subset and aggregation family \\
Heterogeneous experts & One task-aligned expert added at a time using task-specific train or validation data & Evaluate heterogeneous pools and standalone experts \\
\bottomrule
\end{tabular}
\caption{Selection procedures for the main experiment groups. The 6-judge architecture selected on the original 10 cells is reused for the HelpSteer2 validation, FLASK, WildBench, and heterogeneous evaluations.}
\label{tab:appendix-selection-accounting}
\end{table*}

\begin{table*}[ht]
\centering
\scriptsize
\resizebox{.7\textwidth}{!}{
\begin{tabular}{@{}lccccc@{}}
\toprule
Cell & JudgeMoE & Single & Cov-log & $\Delta$ single & $\Delta$ cov-log \\
\midrule
HelpSteer2 val.@1--5 & 0.4394 & 0.3851 & 0.4212 & +0.0543 & +0.0182 \\
HelpSteer2 val.@1--10 & 0.4562 & 0.3805 & 0.4358 & +0.0756 & +0.0204 \\
\midrule
12-cell avg. & 0.3940 & 0.3483 & 0.3552 & +0.0457 & +0.0388 \\
\bottomrule
\end{tabular}}
\caption{Held-out HelpSteer2-validation extension of the selected mainline checkpoint. These two cells were added only after the original 10-cell architecture sweep, so they evaluate the frozen mainline checkpoint without rerunning architecture search. Unlike the FLASK support check, both new HelpSteer2-validation cells are positive against the best single judge and covariance-aware log pooling. The table reports the two new cells together with the resulting 12-cell averages; the corresponding one-sided paired Wilcoxon tests are $p{=}0.0212$ against the best single judge and $p{=}0.0461$ against covariance-aware log pooling.}
\label{tab:appendix-helpsteer2val-mainline-extension}
\end{table*}

\begin{table*}[ht]
\centering
\scriptsize
\resizebox{.7\textwidth}{!}{
\begin{tabular}{@{}lccccc@{}}
\toprule
Cell & JudgeMoE & Single & Cov-log & $\Delta$ single & $\Delta$ cov-log \\
\midrule
FLASK@1--5 & 0.5452 & 0.4897 & 0.5452 & +0.0555 & -0.0001 \\
FLASK@1--10 & 0.4364 & 0.4148 & 0.4449 & +0.0216 & -0.0085 \\
\midrule
12-cell avg. & 0.4011 & 0.3598 & 0.3663 & +0.0413 & +0.0349 \\
\bottomrule
\end{tabular}}
\caption{Secondary FLASK support check for the selected mainline checkpoint. These two cells were added only after the original 10-cell architecture sweep, so they evaluate the frozen mainline checkpoint without rerunning architecture search. FLASK@1--5 ties covariance-aware log pooling up to rounding, consistent with the mixed fixed-pool picture. The table reports the two new FLASK cells together with the resulting 12-cell averages; the corresponding one-sided paired Wilcoxon tests are $p{=}0.0261$ against the best single judge and $p{=}0.1018$ against covariance-aware log pooling. We use FLASK mainly as out-of-domain support for the best-single trend; HelpSteer2 validation remains the primary fixed-pool expansion check.}
\label{tab:appendix-flask-mainline-extension}
\end{table*}

\begin{table*}[t]
\centering
\footnotesize
\setlength{\tabcolsep}{4pt}
\begin{tabular}{@{}llccclc@{}}
\toprule
Setting & Comparison & Left & Right & $\Delta$ & 95\% CI & $p$ \\
\midrule
SummEval 1--10 + BERTScore & Het. JudgeMoE vs. best single & 0.6703 & 0.5172 & +0.1531 & [0.1104, 0.1944] & 0.0007 \\
SummEval 1--10 + BERTScore & Het. JudgeMoE vs. uniform log & 0.6703 & 0.1930 & +0.4773 & [0.4219, 0.5320] & 0.0007 \\
SummEval 1--10 + BERTScore & Het. JudgeMoE vs. hom. JudgeMoE & 0.6703 & 0.6460 & +0.0242 & [-0.0026, 0.0509] & 0.0733 \\
\midrule
HelpSteer2 1--10 + reward & Het. JudgeMoE vs. best single & 0.5454 & 0.3900 & +0.1553 & [0.1468, 0.1644] & 0.0025 \\
HelpSteer2 1--10 + reward & Het. JudgeMoE vs. uniform log & 0.5454 & 0.3102 & +0.2352 & [0.2239, 0.2472] & 0.0025 \\
HelpSteer2 1--10 + reward & Het. JudgeMoE vs. hom. JudgeMoE & 0.5454 & 0.5330 & +0.0124 & [0.0066, 0.0189] & 0.0025 \\
\bottomrule
\end{tabular}
\caption{Paired bootstrap significance for the strongest heterogeneous follow-ups. ``Left'' and ``Right'' are Spearman correlations; $\Delta$ is left minus right. Intervals and $p$-values come from paired bootstrap over matched examples.}
\label{tab:appendix-heterogeneous-significance}
\end{table*}

\begin{table*}[ht]
\centering
\scriptsize
\resizebox{.8\textwidth}{!}{
\begin{tabular}{@{}p{4cm}cccc@{}}
\toprule
Pool & Best single & Uniform log & Cov-log & JudgeMoE \\
\midrule
3-judge LLM pool & 0.1898 & 0.2031 & 0.1988 & \best{0.2468} \\
3-judge LLM + COMET & \best{0.4326} & 0.3926 & 0.3794 & 0.1620 \\
\bottomrule
\end{tabular}}
\caption{WMT MQM@1--10 follow-up with COMET as a task-aligned heterogeneous expert. COMET is a much stronger standalone evaluator than any constituent LLM judge, but appending it to the mixed pool degrades JudgeMoE sharply.}
\label{tab:appendix-comet-translation-followup}
\end{table*}

\begin{table*}[ht]
\centering
\scriptsize
\resizebox{\textwidth}{!}{%
\begin{tabular}{@{}p{3cm}cccccc@{}}
\toprule
Setting & Best single & Full JudgeMoE & $\Delta$ vs.\ single & min frac $>$ single & min frac $\geq 90\%$ gain & min frac $\geq 95\%$ gain \\
\midrule
SummEval 1--10 + BERTScore (repaired) & 0.5172 & 0.6682 & 0.1510 & 0.05 & 0.40 & 1.00 \\
HelpSteer2 1--10 + reward & 0.3900 & 0.5670 & 0.1769 & 0.05 & 0.70 & 0.70 \\
\bottomrule
\end{tabular}
}
\caption{Sample-efficiency summary for the two strongest heterogeneous positives. Fractions refer to the portion of the training split used to fit \methodshort; the gain-recovery columns normalize against the full-data gain over the best single judge. Both settings beat the best single judge with only 5\% of training data, but the reward-model append saturates more slowly than the BERTScore expert.}
\label{tab:appendix-sample-efficiency-followups}
\end{table*}

\begin{table*}[t]
\centering
\scriptsize
\resizebox{\textwidth}{!}{%
\begin{tabular}{@{}p{3cm}p{2.0cm}ccc p{1.5cm}@{}}
\toprule
Setting & Expert & mean $\Delta$ vs.\ hom. abs.-err. & mean $\Delta$ vs.\ single abs.-err. & top-gate rate & Role \\
\midrule
SummEval 1--10 + BERTScore & BERTScore & 0.0744 & 0.3167 & 0.1203 & frequent expert \\
HelpSteer2 1--10 + reward & Skywork reward & 0.0031 & 0.0321 & 0.0000 & soft calibration \\
\bottomrule
\end{tabular}
}
\caption{Compact qualitative summary for the strongest heterogeneous positives. The top-gate rate measures how often the added expert becomes the highest-weight judge on a test example. BERTScore acts like a genuine conditional expert on SummEval, whereas the reward model on HelpSteer2 behaves more like a soft calibration signal.}
\label{tab:appendix-qualitative-expert-behavior}
\end{table*}

\begin{table*}[ht]
\centering
\scriptsize
\begin{tabular}{@{}p{3cm}p{1.6cm}ccccc@{}}
\toprule
Setting & Strong single & Best single & Hom.\ JudgeMoE & Het.\ JudgeMoE & $\Delta$ Het. & $\Delta$ Hom. \\
\midrule
SummEval@1--10 & Qwen2.5-72B & 0.5172 & 0.6460 & 0.6703 & 0.1779 & 0.1536 \\
HelpSteer2@1--10 & Qwen2.5-72B & 0.3900 & 0.5330 & 0.5454 & 0.0716 & 0.0593 \\
WildBench@1--10 & Qwen2.5-72B & 0.3520 & 0.1719 & 0.0854 & 0.0822 & 0.1688 \\
\bottomrule
\end{tabular}
\caption{Aligned strong single-judge follow-ups using Qwen2.5-72B-Instruct on the same evaluation splits as the heterogeneous pilots. JudgeMoE remains stronger on the two positive follow-up cells, while WildBench remains difficult for all methods.}
\label{tab:appendix-strong-single-followups}
\end{table*}

\begin{table*}[ht]
\centering
\scriptsize
\begin{tabular}{@{}p{3cm}rrrrrrr@{}}
\toprule
Setting & Groups & Avg.\ cand. & Best group score & Het. & Hom. & Single & $\Delta$ Het. \\
\midrule
summeval best-of-$N$ & 100 & 9.3200 & 0.9979 & 0.9365 & 0.9446 & 0.9194 & -0.0081 \\
\bottomrule
\end{tabular}
\caption{Offline best-of-$N$ SummEval case study, grouping candidate summaries by source article. The repaired heterogeneous pool still beats the best single judge, while the repaired homogeneous pool remains stronger on this reranking-style proxy.}
\label{tab:appendix-downstream-case-study}
\end{table*}

\subsection{Implementation Details}

Table~\ref{tab:appendix-implementation} summarizes the implementation choices that matter for reproducibility.

\begin{table*}[ht]
\centering
\footnotesize
\begin{tabular}{@{}lp{7cm}@{}}
\toprule
Component & Setting \\
\midrule
Direct-score target & Human scores normalized to $[0,1]$ \\
Direct loss & Huber loss on the fused score \\
Pairwise target & Binary preference converted to $\{-1,+1\}$ \\
Pairwise loss & Margin-ranking loss with tuned auxiliary weight and fixed margin $m{=}0.0$ \\
Large sweeps & JSONL-based resume support for interrupted runs \\
Subset comparisons & Exact example-ID overlap for sample-aligned evaluation \\
\bottomrule
\end{tabular}
\caption{Implementation details relevant to reproduction.}
\label{tab:appendix-implementation}
\end{table*}

Selection differs across the paper's components, so Appendix Table~\ref{tab:appendix-selection-accounting} summarizes the main procedures. The protocol study evaluates scoring choices separately from aggregation. The main aggregation benchmark uses one shared 6-judge configuration across the original 10 direct-score cells, with 1--5 as the default scale and 1--10 as the main alternative. The six additional cells from HelpSteer2 validation, FLASK, and WildBench were evaluated after the judge pool, architecture, feature construction, and training procedure had been fixed. The compact 3-judge pool is selected once from the protocol-study screening and then held fixed. Appendix Table~\ref{tab:pairwise-main} reports a separate 3-judge pairwise experiment. For the WMT subset experiment, judge subsets are selected on WMT validation data and then retrained and evaluated on the corresponding test split.

\begin{figure}[t]
\centering
\includegraphics[width=\columnwidth]{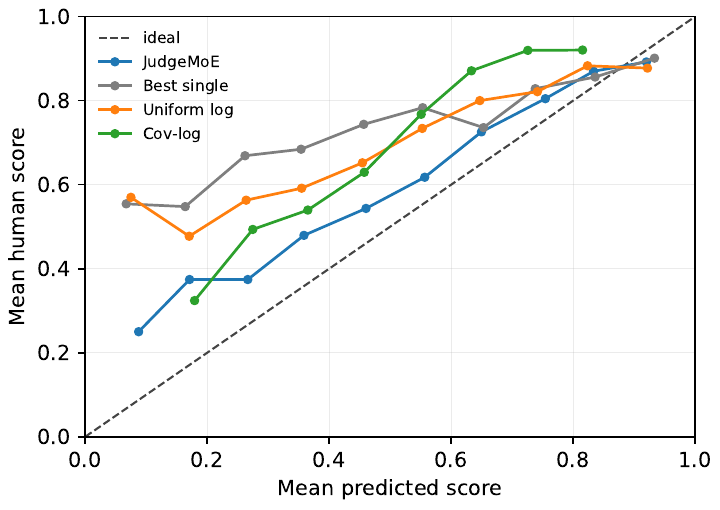}
\caption{Equal-cell reliability diagram for the main direct-score benchmark. For each system and each bin, we average the bin-wise predicted score and human score across the 10 benchmark cells, so the plot complements the equal-cell ECE table and the example-weighted summary.}
\label{fig:appendix-calibration-reliability}
\end{figure}

The revision-only interpretability and calibration diagnostics are generated directly from the released per-example artifacts by the diagnostic-table generation program in the artifact packet. That program regenerates Appendix Tables~\ref{tab:appendix-calibration}, \ref{tab:appendix-calibration-by-dataset}, \ref{tab:appendix-gate-cases}, \ref{tab:appendix-wmt-subset-stability}, and~\ref{tab:appendix-wmt-gate-behavior} together with Figure~\ref{fig:appendix-calibration-reliability}.

\subsection{Seed Stability}

Table~\ref{tab:appendix-seeds} summarizes seed stability for the two main settings.

\begin{table}[t]
\centering
\footnotesize
\setlength{\tabcolsep}{4pt}
\begin{tabular}{@{}lccc@{}}
\toprule
Setting & Mean Sp. & Std. & $\Delta$ vs. single \\
\midrule
Mainline 6-judge & 0.3826 & 0.0013 & 0.0413 \\
Efficient 3-judge & 0.3638 & 0.0028 & 0.0245 \\
\bottomrule
\end{tabular}
\caption{Seed stability for the strongest 6-judge and 3-judge settings. The small standard deviations show that the two reported deployment settings are stable across random seeds.}
\label{tab:appendix-seeds}
\end{table}

\subsection{Pairwise Joint-Training Details}

Table~\ref{tab:appendix-pairwise-breakdown} gives the per-dataset pairwise improvements for the best joint setting (pairwise loss weight 0.5). These results support the main claim that pairwise supervision is best used as an auxiliary signal alongside direct-score supervision.

\begin{table}[t]
\centering
\footnotesize
\setlength{\tabcolsep}{4pt}
\begin{tabular}{@{}lcc@{}}
\toprule
$\lambda$ & Direct Sp. & $\Delta$ pair acc. \\
\midrule
0.1 & 0.3725 & 0.0360 \\
0.5 & \best{0.4192} & \best{0.0462} \\
1.0 & 0.4190 & 0.0431 \\
\bottomrule
\end{tabular}
\caption{Joint training with pairwise auxiliary loss in the separate 3-judge pairwise-ablation setting (MT-Bench-human and Summarize-from-Feedback validation, each at 1--5 and 1--10). The direct Spearman values summarize this 4-row setting, distinct from the 10-cell 6-judge mainline benchmark. Moderate pairwise weight improves pairwise accuracy without hurting direct-score quality.}
\label{tab:pairwise-main}
\end{table}

\begin{table}[t]
\centering
\footnotesize
\setlength{\tabcolsep}{4pt}
\begin{tabular}{@{}lp{2.15cm}cc@{}}
\toprule
Scale & Dataset & Base acc. & Joint acc. \\
\midrule
1--5 & MT-Bench-human & 0.6369 & 0.6823 \\
1--5 & Summarize-from-Feedback & 0.5140 & 0.6054 \\
1--10 & MT-Bench-human & 0.7200 & 0.7250 \\
1--10 & Summarize-from-Feedback & 0.5994 & 0.6244 \\
\bottomrule
\end{tabular}
\caption{Per-dataset pairwise gains for the best joint-training setting. The joint model improves on both pairwise datasets at both scales, which is why the paper treats pairwise supervision as a useful auxiliary signal for direct-score training.}
\label{tab:appendix-pairwise-breakdown}
\end{table}

\begin{table}[t]
\centering
\scriptsize
\setlength{\tabcolsep}{3pt}
\begin{tabular}{@{}lccccc@{}}
\toprule
Scale & Best single & Uniform log & Direct & Joint & Pair-only \\
\midrule
1--5  & 0.5482 & 0.5385 & 0.5411 & \textbf{0.5495} & 0.5112 \\
1--10 & \textbf{0.5674} & 0.5578 & 0.5601 & 0.5507 & 0.5026 \\
\bottomrule
\end{tabular}
\caption{Supplementary OOD evaluation on 3{,}114 English non-tie Arena human preferences. The pattern is consistent with the main pairwise story: joint training helps at 1--5, pairwise-only remains weak at both scales, and at 1--10 the strongest single judge is already difficult to improve upon.}
\label{tab:appendix-arena-pairwise}
\end{table}

\begin{table}[t]
\centering
\scriptsize
\setlength{\tabcolsep}{3pt}
\begin{tabular}{@{}lccccc@{}}
\toprule
Scale & Best single & Uniform log & Direct & Joint & Pair-only \\
\midrule
1--5  & 0.4870 & 0.4479 & \textbf{0.4917} & 0.4656 & 0.4562 \\
1--10 & \textbf{0.5201} & 0.4870 & 0.5036 & 0.5012 & 0.5130 \\
\bottomrule
\end{tabular}
\caption{Supplementary OOD evaluation on a more recent, length-controlled Arena Expert subset with 844 English A/B comparisons. The benchmark is filtered to contexts that fit the local three-judge pool without OOM failures (`source $\leq$ 3000` chars; responses $\leq$ 6000 chars). The result is more conservative than Arena Human Preference: direct-only is the strongest learned setting at 1--5, while at 1--10 the best single judge remains strongest.}
\label{tab:appendix-arena-expert-pairwise}
\end{table}

\begin{table}[t]
\centering
\scriptsize
\setlength{\tabcolsep}{3pt}
\begin{tabular}{@{}lccccc@{}}
\toprule
Scale & Best single & Uniform log & Direct & Joint & Pair-only \\
\midrule
1--5  & \textbf{0.5605} & 0.5446 & 0.5361 & 0.5259 & 0.4952 \\
1--10 & \textbf{0.5821} & 0.5685 & \textbf{0.5821} & 0.5657 & 0.4832 \\
\bottomrule
\end{tabular}
\caption{Supplementary OOD evaluation on a recent length-controlled PPE-Human-Preference-V1 subset with 1,759 English A/B comparisons. The benchmark is filtered to `source $\leq$ 4000` chars and responses `$\leq$ 8000` chars. The result is more conservative than Arena Human Preference: at 1--5 the best single judge remains strongest, and at 1--10 direct-only only matches the best single judge while joint training no longer helps.}
\label{tab:appendix-ppe-pairwise}
\end{table}

\begin{table}[t]
\centering
\scriptsize
\setlength{\tabcolsep}{3pt}
\begin{tabular}{@{}lccccc@{}}
\toprule
Scale & Best single & Uniform log & Cov-log & JudgeMoE & Hard \\
\midrule
1--5  & 0.2285 & 0.2183 & \textbf{0.2638} & 0.1955 & 0.0804 \\
1--10 & 0.2351 & 0.2650 & 0.2589 & \textbf{0.2709} & -0.0157 \\
\bottomrule
\end{tabular}
\caption{Supplementary direct-score evaluation on a modern open-ended WildBench v2-hard subset. We use the released train split, take the provided GPT-4 reference response as the candidate answer, and keep a length-controlled subset of 246 examples (`prompt $\leq$ 6000` chars; candidate `$\leq$ 8000` chars). The pattern remains conditional: covariance-aware log pooling is strongest at 1--5, while the 6-judge \methodshort mainline becomes strongest at 1--10.}
\label{tab:appendix-wildbench-direct}
\end{table}

\begin{table}[t]
\centering
\small
\setlength{\tabcolsep}{5pt}
\begin{tabular}{@{}lrrrr@{}}
\toprule
Standalone evaluator & Cells & Standalone & JudgeMoE & JMoE wins \\
\midrule
DeepSeek-V3.2 & 4 & 0.5710 & 0.4966 & 1/4 \\
Qwen2.5-72B & 4 & 0.5362 & 0.4966 & 1/4 \\
\bottomrule
\end{tabular}
\caption{Aligned comparisons with larger standalone evaluators on the same evaluation subsets. These evaluators are included as standalone deployment options in the validation analysis.}
\label{tab:large-judge-smallset}
\end{table}

\begin{table*}[t]
\centering
\footnotesize
\setlength{\tabcolsep}{4pt}
\begin{tabular}{@{}p{0.30\linewidth}p{0.34\linewidth}p{0.18\linewidth}@{}}
\toprule
System & Active params / token & Relative compute \\
\midrule
Local single judge & 7--12B dense & 0.19--0.32$\times$ \\
Local 3-judge efficient pool & 28B dense across separate passes & 0.76$\times$ \\
Local 6-judge mainline pool & 53B dense across separate passes & 1.43$\times$ \\
DeepSeek-V3.2-Exp & 37B activated, 671B total & 1.00$\times$ \\
\bottomrule
\end{tabular}
\caption{Hardware-independent first-order inference-compute proxy. For autoregressive inference, weight-matrix FLOPs scale approximately linearly with active parameters and processed tokens, so under the same evaluation workload the relative compute is well approximated by active-parameter counts per token. Under the 1K-example SummEval 1--5 workload used in our deployment note (roughly 474K processed tokens), this proxy corresponds to about 26.5 PFLOPs for the local 3-judge pool, 50.2 PFLOPs for the local 6-judge pool, and 35.1 PFLOPs for DeepSeek-V3.2-Exp. The six local judges are independent and can be run in parallel; in the same probe, single-judge jobs took 16.8--27.4 minutes, so ideal six-way parallel latency is bounded by the slowest job. The sequential sum was 134.4 minutes.}
\label{tab:appendix-cost-benchmark}
\end{table*}

\begin{table*}[t]
\centering
\scriptsize
\setlength{\tabcolsep}{3pt}
\begin{tabular}{@{}cp{0.14\linewidth}p{0.13\linewidth}ccccp{0.17\linewidth}p{0.17\linewidth}@{}}
\toprule
Type & Cell & Example & Human & JudgeMoE & Cov-log & Top gate & Judge scores & Gate weights \\
\midrule
+ & summeval@1-5 & \texttt{S1499-relevance} & 0.833 & 0.825 & 0.344 & L3 & [.11,1,.50,.32,.01,.43] & [.08,.25,.35,.15,.04,.13] \\
+ & helpsteer2@1-10 & \texttt{H10914-helpfulness} & 1.000 & 0.901 & 0.524 & G3 & [.57,.90,.52,.50,.50,.55] & [0,.98,0,.01,0,0] \\
+ & wmt-mqm@1-10 & \texttt{W66502} & 0.800 & 0.798 & 0.564 & G2 & [.77,.64,.54,.57,.52,.56] & [1,0,0,0,0,0] \\
+ & roscoe-esnli@1-10 & \texttt{R30} & 1.000 & 0.822 & 0.580 & G3 & [.56,.90,.50,.54,.49,.60] & [.18,.41,0,.17,0,.23] \\
- & topical-chat@1-5 & \texttt{T12-Engaging} & 0.333 & 0.898 & 0.390 & G3 & [.39,1,.51,.39,.42,.34] & [.10,.31,.15,.03,.16,.24] \\
\bottomrule
\end{tabular}
\caption{Illustrative per-example cases where learned aggregation improves over (+) or fails against (-) fixed cov-log pooling. Judge-score and gate-weight vectors use the order G2/G3/L3/Min/Q2.5/Q3.5. Examples are selected automatically by the largest absolute-error change in the indicated direction. Across the 10 direct-score cells, the JudgeMoE prediction is closer to the human score than cov-log in 75.5\% of examples on an equal-cell basis (84.6\% example-weighted).}
\label{tab:appendix-gate-cases}
\end{table*}

\begin{table}[t]
\centering
\footnotesize
\setlength{\tabcolsep}{4pt}
\begin{tabular}{@{}lccccc@{}}
\toprule
Scale & Cand. & $\rho_s$(valid,test) & Sel. valid & Sel. test & Test rank \\
\midrule
1-5 & 120 & 0.973 & 0.143 & 0.153 & 1 \\
1-10 & 120 & 0.974 & 0.213 & 0.232 & 3 \\
\bottomrule
\end{tabular}
\caption{Validation--test stability for the WMT sparse fixed-pool search. Candidates include all single judges and all non-single subsets under uniform-log and cov-log pooling. The selected candidate is chosen by validation Spearman; its test rank is computed among the same candidates by test Spearman.}
\label{tab:appendix-wmt-subset-stability}
\end{table}

\begin{table*}[t]
\centering
\footnotesize
\setlength{\tabcolsep}{4pt}
\begin{tabular}{@{}llccccp{0.30\linewidth}@{}}
\toprule
Scale & Judge & Judge Sp. & MAE & Mean gate & Top-gate & Diagnostic readout \\
\midrule
1-5 & G2 & 0.141 & 0.186 & 0.373 & 46.1\% & best single; largest gate mass \\
1-5 & G3 & 0.019 & 0.256 & 0.244 & 21.7\% & secondary evidence \\
1-5 & L3 & -0.030 & 0.269 & 0.221 & 19.2\% & secondary evidence \\
1-5 & Min & 0.086 & 0.299 & 0.140 & 12.7\% & secondary evidence \\
1-5 & Q2.5 & -0.148 & 0.651 & 0.006 & 0.3\% & known contaminant; gate almost removes it \\
1-5 & Q3.5 & 0.014 & 0.400 & 0.017 & 0.0\% & secondary evidence \\
\midrule
1-10 & G2 & 0.180 & 0.111 & 0.529 & 69.5\% & best single; dominant route \\
1-10 & G3 & 0.128 & 0.122 & 0.120 & 3.8\% & secondary evidence \\
1-10 & L3 & -0.086 & 0.319 & 0.000 & 0.0\% & negative single; gate suppresses \\
1-10 & Min & 0.122 & 0.163 & 0.117 & 12.4\% & secondary evidence \\
1-10 & Q2.5 & 0.037 & 0.292 & 0.054 & 0.5\% & secondary evidence \\
1-10 & Q3.5 & 0.139 & 0.195 & 0.179 & 13.8\% & secondary evidence \\
\bottomrule
\end{tabular}
\caption{WMT per-judge gate behavior connecting sparse-repair theory to the learned gate. Judge Spearman/MAE provide an empirical precision proxy; mean gate and top-gate rates are computed from the learned full-pool gate. Rank correlation between per-judge Spearman and mean gate is 1-5: 0.71, 1-10: 1.00. These correlations use only six judges and are diagnostic. The gate identifies weak judges, but at 1--5 sparse retraining is still needed because coarse scores leave too little coherent signal in the contaminated full pool.}
\label{tab:appendix-wmt-gate-behavior}
\end{table*}

\begin{table}[t]
\centering
\small
\setlength{\tabcolsep}{4pt}
\begin{tabular}{@{}llc@{}}
\toprule
Scale / subset & System or diagnostic & Spearman \\
\midrule
1--5 (full) & Best single judge & 0.1413 \\
1--5 (full) & Top-2 sparse log pool & 0.1473 \\
1--5 (full) & Sparse JudgeMoE (2 selected judges) & \best{0.1648} \\
1--5 (full) & Full-6 uniform log pool & 0.0016 \\
1--5 (full) & Mainline \methodshort & 0.0376 \\
\midrule
1--10 (full) & Best single judge & 0.1797 \\
1--10 (full) & Top-4 sparse log pool & 0.2161 \\
1--10 (full) & Sparse JudgeMoE (4 selected judges) & 0.2246 \\
1--10 (full) & Full-6 uniform log pool & 0.1504 \\
1--10 (full) & Mainline \methodshort & \best{0.2293} \\
\bottomrule
\end{tabular}
\caption{Deeper WMT MQM failure-and-recovery analysis. At 1--5, naive all-judge fusion degrades, but held-out-validation-selected sparse composition recovers the signal and a sparse JudgeMoE trained on the selected subset performs best. At 1--10, better score resolution already makes the full 6-judge JudgeMoE competitive, while the sparse 4-judge JudgeMoE retains most of that performance with a cleaner pool.}
\label{tab:wmt-failure}
\end{table}

\subsection{Additional WMT Diagnostic Visualization}

\begin{figure}[t]
\centering
\includegraphics[width=\columnwidth]{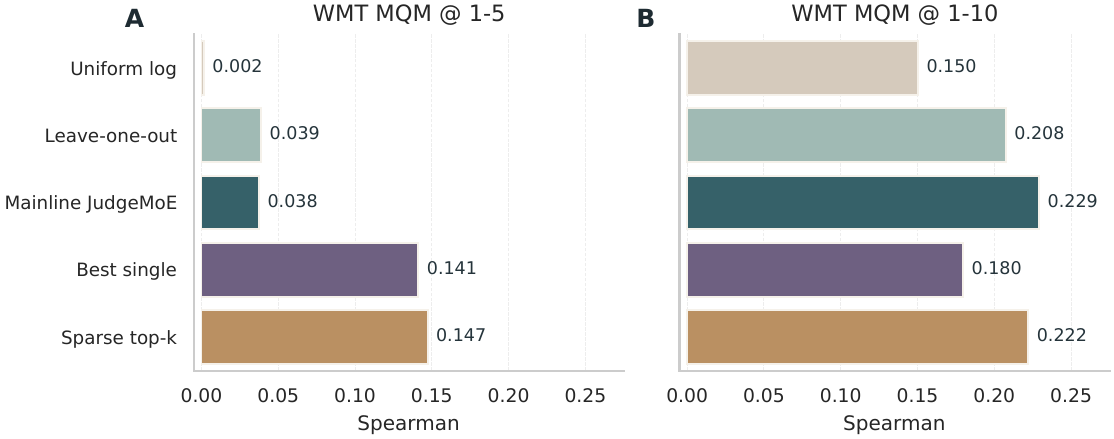}
\caption{Compact visualization of the WMT MQM failure mode. At 1--5, sparse composition and dropping one weak judge both improve over naive full-pool aggregation, and the sparse retraining follow-up shows that a cleaned judge pool can repair the failure. At 1--10, better score resolution lifts both sparse pooling and the learned mainline model, with sparse retraining retaining most of the mainline gain.}
\label{fig:appendix-wmt-failure}
\end{figure}

\subsection{Focused Theoretical Perspective}

This section keeps only the theory that ties directly to the JudgeMoE formulation and the empirical storyline. The propositions below explain three concrete design and deployment choices: why score distributions can contain more information than compressed scores, why the best pooling temperature is often $T{<}1$, and when sparse repair helps a contaminated log pool. Several small judges can still lose to a larger judge, as the WMT failure case shows. Proposition~1 together with Propositions~3 and~4 are retained as background facts; Propositions~6--8 are specific to the JudgeMoE setup and connect directly to the main empirical findings.

\paragraph{Notation.}
Let the ordered human score take values in $\{v_1,\dots,v_K\}\subset\mathbb{R}$. A judge outputs a categorical distribution $P=(P_1,\dots,P_K)$ over these values. For multi-judge pooling, let $w\in\Delta^{m-1}$ denote a pooling weight vector. When comparing two candidate score buckets $a$ and $b$, write the per-judge likelihood ratio as $r_j(a,b)=p_j(a)/p_j(b)$.

\paragraph{Proposition 1 (Conditional mean minimizes squared loss).} Let the ordered human score take values in $\{v_1,\dots,v_K\}\subset\mathbb{R}$, let a judge output a categorical distribution $P=(P_1,\dots,P_K)$ over these values, and define $m(P)=\mathbb{E}[Y\mid P]$. Then for any measurable scalar summary $a(P)$,
\begin{equation}
\mathbb{E}\!\left[(Y-a(P))^2\mid P\right]=\mathrm{Var}(Y\mid P)+\bigl(m(P)-a(P)\bigr)^2.
\end{equation}
Hence the unique minimizer of conditional squared loss is
\begin{equation}
a^\star(P)=m(P).
\end{equation}
If the judge distribution is calibrated with respect to the human score distribution, so that
\begin{equation}
m(P)=\sum_{k=1}^{K}v_kP_k\equiv\mu(P),
\end{equation}
then the soft score $\mu(P)$ is the optimal scalar summary under squared loss. Under the same calibration condition, the hard score $h(P)=v_{\arg\max_k P_k}$ is a Bayes action under exact-class $0$--$1$ loss.

\paragraph{Proof sketch.} Apply the bias--variance decomposition around the conditional mean $m(P)=\mathbb{E}[Y\mid P]$. The variance term is independent of $a(P)$, so the conditional squared loss is minimized at $a(P)=m(P)$. Under score-distribution calibration, $m(P)=\mu(P)$.

\paragraph{Interpretation.} Expected scoring is the natural one-dimensional reduction when the judge distribution is aligned with the human-score distribution. The empirical protocol study examines how well this reduction works with extracted LLM score probabilities in practice.

\paragraph{Proposition 3 (Correlated error floor for averaging).}
Assume judge score errors $\varepsilon_j$ are zero mean with
\[
\mathrm{Var}(\varepsilon_j)=\sigma^2, \qquad
\mathrm{Corr}(\varepsilon_i,\varepsilon_j)=\rho \ \text{for}\ i\neq j,
\]
where $\rho \in [0,1)$. For uniform averaging,
\[
\bar{\varepsilon} = \frac{1}{m}\sum_{j=1}^{m}\varepsilon_j,
\]
the error variance is
\begin{equation}
\mathrm{Var}(\bar{\varepsilon})
=
\sigma^2\left(\rho + \frac{1-\rho}{m}\right).
\end{equation}
Therefore,
\begin{equation}
\lim_{m \to \infty}\mathrm{Var}(\bar{\varepsilon}) = \sigma^2 \rho.
\end{equation}

\paragraph{Proof sketch.}
Expand $\mathrm{Var}(\bar{\varepsilon})$ into diagonal and off-diagonal covariance terms. Under the exchangeable-correlation assumption, the diagonal terms contribute $\sigma^2/m$ and the off-diagonal terms contribute $\sigma^2\rho(1-1/m)$.

\paragraph{Interpretation.}
Ensembling removes only the idiosyncratic component of judge noise. Shared bias or highly correlated error remains as an irreducible floor, which formalizes why judge diversity matters more than simply adding more judges.

\paragraph{Proposition 4 (Weak-judge veto threshold under log pooling).}
Consider a weighted log pool
\begin{equation}
\begin{aligned}
q(y)&\propto \exp\!\left(\sum_{j=1}^{m}\alpha_j \log p_j(y)\right), \\
\alpha_j&\ge 0,\qquad \sum_j \alpha_j=1.
\end{aligned}
\end{equation}
For two competing score buckets $a$ and $b$,
\begin{equation}
\log\frac{q(a)}{q(b)}
=
\sum_{j=1}^{m}\alpha_j \log\frac{p_j(a)}{p_j(b)}.
\end{equation}
Hence, if one weak judge $j=0$ prefers $b$ over $a$ while the remaining judges prefer $a$ over $b$, the pooled ranking flips in favor of $b$ if and only if
\begin{equation}
\alpha_0 \log\frac{p_0(b)}{p_0(a)}
>
\sum_{j\neq 0}\alpha_j \log\frac{p_j(a)}{p_j(b)}.
\end{equation}
In the homogeneous equal-weight case with $m$ good judges, $p_j(a)=p$, $p_j(b)=p'$, and one weak judge with $p_0(a)=\delta$, $p_0(b)=\delta'$, this reduces to
\begin{equation}
\log(\delta'/\delta) > m \log(p/p').
\end{equation}

\paragraph{Proof sketch.}
Take the log ratio of the pooled probabilities for $a$ and $b$; the result is exactly the weighted sum of per-judge log-likelihood ratios. The flip condition is simply the sign change of this quantity.

\paragraph{Interpretation.}
Log pooling rewards agreement, but it also gives every judge multiplicative influence. A weak judge that places too little mass on the correct bucket can therefore veto the remaining judges. This is the mechanism behind the WMT weak-judge pollution result.

\paragraph{Linear-pool contrast.}
Under equal-weight linear pooling in the same homogeneous setting, the ranking flips only when
\[
\delta' - \delta > m (p-p').
\]
The weak judge enters additively in probability space, so single-judge pollution is less explosive than in log pooling.

\paragraph{Proposition 6 (Optimal pooling temperature as a calibration diagnostic).}
Suppose the ground-truth posterior over $K$ ordered scores given instance $x$ is $p^{*}(\cdot \mid x)$. Model each judge $j$ as producing a temperature-scaled version of that posterior:
\begin{equation}\label{eq:cal-model}
  p_k^{(j)}(x)
  =
  \frac{\bigl(p_k^{*}(x)\bigr)^{1/\tau_j}}{Z_j(x)},
  \qquad
  Z_j(x) = \sum_{\ell=1}^{K}\bigl(p_\ell^{*}(x)\bigr)^{1/\tau_j},
\end{equation}
where $\tau_j > 0$ is the calibration temperature of judge~$j$ ($\tau_j = 1$ calibrated, $\tau_j > 1$ under-confident, $\tau_j < 1$ over-confident). Under equal-weight tempered log pooling at temperature $T$, the fused distribution satisfies
\begin{equation}\label{eq:gamma}
  q_k^{T}(x) \propto \bigl(p_k^{*}(x)\bigr)^{\gamma(T)},
  \qquad
  \gamma(T) = \frac{1}{T\,\bar{\tau}_H},
\end{equation}
where $\bar{\tau}_H = m \big/ \sum_{j=1}^{m}\tau_j^{-1}$ is the harmonic mean of the judges' calibration temperatures. The fused distribution recovers the true posterior ($q^T = p^*$) if and only if
\begin{equation}\label{eq:Tstar}
  T^{*} = \frac{1}{\bar{\tau}_H}
       = \frac{1}{m}\sum_{j=1}^{m}\frac{1}{\tau_j}.
\end{equation}
At this temperature the expected-score predictor $\hat{y}^{T^*}\!(x) = \sum_k s_k\,q_k^{T^*}\!(x)$ equals the Bayes-optimal point prediction $\mathbb{E}[y \mid x]$, and for any $T \neq T^*$ the conditional MSE increases by the squared bias:
\begin{equation}\label{eq:T-mse}
  \mathbb{E}\bigl[(\hat{y}^{T} - y)^2 \mid x\bigr]
  =
  \operatorname{Var}(y \mid x)
  +
  \bigl(\hat{y}^{T}(x) - \hat{y}^{T^*}\!(x)\bigr)^2.
\end{equation}

\paragraph{Proof sketch.}
The equal-weight tempered log pool gives
\[
  \log q_k^{T}(x)
  = \frac{1}{T}\cdot\frac{1}{m}\sum_{j=1}^{m}\log p_k^{(j)}(x)
    + \mathrm{const.}
\]
Substituting \eqref{eq:cal-model},
\[
  \log p_k^{(j)}(x)
  = \tau_j^{-1}\log p_k^{*}(x) - \log Z_j(x).
\]
The normalization terms are constant across $k$ and vanish into the pooled normalizer, so
\[
  \log q_k^{T}(x)
  =
  \underbrace{\frac{1}{mT}\sum_{j=1}^{m}\frac{1}{\tau_j}}_{\gamma(T)}
  \log p_k^{*}(x) + \mathrm{const.}
\]
Writing $\sum_j \tau_j^{-1} = m/\bar{\tau}_H$ yields \eqref{eq:gamma}; setting $\gamma(T)=1$ gives $T^* = 1/\bar{\tau}_H$. The MSE identity then follows from Proposition~1 and the standard bias--variance decomposition around $\mathbb{E}[y \mid x]$.

\paragraph{Interpretation.}
Proposition~6 turns the pooling temperature into a calibration diagnostic as well as an empirical hyperparameter. The empirical preference for $T{=}0.7$ in Table~\ref{tab:temp-sensitivity} is therefore consistent with a collectively under-confident judge pool.

\paragraph{Scope.}
The temperature-scaled model above is a first-order approximation. Real judges exhibit instance-dependent miscalibration, so Proposition~6 characterizes the pool-average calibration condition without predicting one exact optimal $T$ for every example. We use it only as a directional diagnostic for the temperature sweep in Table~\ref{tab:temp-sensitivity}. Its practical use is to explain why the best temperature often lies below one and to motivate initialization of the temperature sweep from a small calibration set.

\paragraph{Proposition 7 (Distribution--score information gap).}
Let $\mathbf{P} = (p^{(1)},\ldots,p^{(m)}) \in (\Delta^{K-1})^m$ be the distributional evidence from $m$ judges over $K \ge 3$ ordered scores, and let $\boldsymbol{\mu} = (\mu_1,\ldots,\mu_m)$ with $\mu_j = \sum_k s_k\,p_k^{(j)}$ be the corresponding score vector of expected scores.

Here $\sigma(\mathbf{P})$ denotes the information contained in the full collection of judge distributions, while $\sigma(\boldsymbol{\mu})$ denotes the coarser information retained after each distribution is reduced to its expected score. Thus $\sigma(\boldsymbol{\mu}) \subseteq \sigma(\mathbf{P})$ simply formalizes the fact that expected scores are deterministic summaries of the full distributions.

\textbf{(i)} The Bayes-optimal predictors $f^*(\boldsymbol{\mu}) = \mathbb{E}[y \mid \boldsymbol{\mu}]$ and $g^*(\mathbf{P}) = \mathbb{E}[y \mid \mathbf{P}]$ satisfy
\begin{equation}\label{eq:info-gap}
  \begin{aligned}
  &\underbrace{\mathbb{E}\bigl[(y - f^*(\boldsymbol{\mu}))^2\bigr]}_{\text{score risk}}
  -
  \underbrace{\mathbb{E}\bigl[(y - g^*(\mathbf{P}))^2\bigr]}_{\text{distribution risk}}
  \\
  &\qquad =
  \mathbb{E}\Bigl[\operatorname{Var}\bigl(g^*(\mathbf{P}) \mid \boldsymbol{\mu}\bigr)\Bigr]
  \ge 0.
  \end{aligned}
\end{equation}

\textbf{(ii)} For any auxiliary feature $\phi(\mathbf{P})$ with information beyond $\boldsymbol{\mu}$, the gap admits the lower bound
\begin{equation}\label{eq:info-lb}
  \Delta_{\mathrm{info}}
  \ge
  \mathbb{E}_{\boldsymbol{\mu}}\!\left[
    \frac{\bigl(\operatorname{Cov}(y,\phi(\mathbf{P}) \mid \boldsymbol{\mu})\bigr)^2}
         {\operatorname{Var}(\phi(\mathbf{P}) \mid \boldsymbol{\mu})}
  \right].
\end{equation}

\textbf{(iii)} When $K = 2$, $\sigma(\boldsymbol{\mu}) = \sigma(\mathbf{P})$ and the gap is exactly zero. When $K \ge 3$ and judge distributions are generically non-degenerate, $\sigma(\boldsymbol{\mu}) \subsetneq \sigma(\mathbf{P})$ and the gap is generically strict.

\paragraph{Proof sketch.}
Part~(i) is the law of total variance applied to $\sigma(\boldsymbol{\mu}) \subseteq \sigma(\mathbf{P})$:
\[
  \begin{aligned}
  \operatorname{Var}(y \mid \boldsymbol{\mu})
  &=
  \mathbb{E}\bigl[\operatorname{Var}(y \mid \mathbf{P}) \mid \boldsymbol{\mu}\bigr]
  \\
  &\quad+
  \operatorname{Var}\bigl(\mathbb{E}[y \mid \mathbf{P}] \mid \boldsymbol{\mu}\bigr).
  \end{aligned}
\]
Taking expectations yields \eqref{eq:info-gap}. For part~(ii), conditional Cauchy--Schwarz gives
\[
  \operatorname{Var}\bigl(g^*(\mathbf{P}) \mid \boldsymbol{\mu}\bigr)
  \ge
  \frac{\bigl(\operatorname{Cov}(g^*(\mathbf{P}),\phi(\mathbf{P}) \mid \boldsymbol{\mu})\bigr)^2}
       {\operatorname{Var}(\phi(\mathbf{P}) \mid \boldsymbol{\mu})},
\]
and the tower property identifies the covariance term with $\operatorname{Cov}(y,\phi(\mathbf{P}) \mid \boldsymbol{\mu})$. Part~(iii) follows because for $K=2$ the simplex is one-dimensional, so the map from distribution to expected score is affine and invertible, whereas for $K \ge 3$ it has a non-trivial kernel inside $\Delta^{K-1}$.

\paragraph{Interpretation.}
Proposition~7 formalizes the paper's central formulation choice. Distributional aggregation has a genuine information advantage whenever judge distributions vary in ways that are unrecoverable from their expected scores alone. This is the setting in which entropy, confidence, and near-tie structure become useful.

\paragraph{Example.}
For scores $\{1,2,3\}$, the distributions $(0.5,0,0.5)$ and $(0,1,0)$ have the same expected score, 2, while encoding very different uncertainty: the first is a polarized near-tie between low and high scores, while the second is concentrated on the middle score. A score-only aggregator treats them the same; a distributional aggregator can distinguish them.

\paragraph{Concrete instantiation.}
Taking $\phi(\mathbf{P})$ to be mean judge entropy, confidence, or maximum-bucket probability yields a computable lower bound on the score-to-distribution gap. Appendix Table~\ref{tab:appendix-feature-ablation} provides the corresponding empirical signal in our setting: moving from distributions alone to the full scalar summary bundle raises average Spearman from 0.3746 to 0.3832, suggesting that these extra distributional statistics carry useful information beyond the expected scores.

\paragraph{Scope.}
The lower bound in \eqref{eq:info-lb} is intentionally general. Its role is diagnostic: it says what kind of auxiliary distributional statistic can certify a non-trivial information gap, while making no claim that any single statistic captures the full gain.

\paragraph{Proposition 8 (Sparse improvement condition for expected-score MSE).}
Under the calibration model \eqref{eq:cal-model} with equal weights and pooling temperature $T$, define judge precision $r_j = \tau_j^{-1}$, mean pool precision $\bar{r} = \frac{1}{m}\sum_{j=1}^{m} r_j$, pool precision excess $\Delta = m(\bar{r} - T)$, and judge precision deviation $\delta_0 = r_0 - T$ for the candidate judge $j_0$ to be dropped. The effective exponent of the full pool is $\gamma_{\mathrm{full}} = \bar{r}/T$, and dropping $j_0$ yields $\gamma_{-0} = (m\bar{r} - r_0)/((m-1)T)$.

Dropping judge $j_0$ moves the effective exponent strictly closer to the optimum $\gamma^* = 1$ (equivalently, $|\gamma_{-0} - 1| < |\gamma_{\mathrm{full}} - 1|$) if and only if
\begin{equation}\label{eq:drop-cond}
  \begin{aligned}
  \delta_0 &\in
  \Bigl(\tfrac{\Delta}{m}, \tfrac{(2m-1)\Delta}{m}\Bigr)
  && \text{when } \Delta > 0,\\
  \delta_0 &\in
  \Bigl(\tfrac{(2m-1)\Delta}{m}, \tfrac{\Delta}{m}\Bigr)
  && \text{when } \Delta < 0.
  \end{aligned}
\end{equation}
Equivalently, $\delta_0$ lies strictly between $\bar{r} - T$ and $(2m-1)(\bar{r} - T)$. When $\Delta = 0$ (the pool is already optimally sharpened), dropping any judge can only hurt or leave the exponent unchanged.

\paragraph{Proof sketch.}
Substitute the effective exponents into the inequality $|\gamma_{-0} - 1| < |\gamma_{\mathrm{full}} - 1|$:
\[
  \gamma_{\mathrm{full}} - 1 = \frac{\Delta}{mT},
  \qquad
  \gamma_{-0} - 1 = \frac{\Delta - \delta_0}{(m-1)T}.
\]
This reduces the problem to
\[
  \frac{|\Delta - \delta_0|}{m-1} < \frac{|\Delta|}{m},
\]
which becomes a quadratic inequality in $\delta_0$. Solving that inequality yields exactly the interval in \eqref{eq:drop-cond}.

\paragraph{Corollary 1 (Weak judges hurt under-sharpened pools).}
When the pool is under-sharpened ($\Delta < 0$), a very weak judge has $r_0 \approx 0$ and therefore $\delta_0 \approx -T$. In this condition, dropping such a judge moves the pooled exponent closer to the optimum and improves expected-score aggregation. This matches the WMT MQM result in Table~\ref{tab:wmt-failure}: removing Qwen2.5-7B raises the full-pool baseline from 0.0016 to 0.0389 at the 1--5 scale.

\paragraph{Corollary 2 (Weak judges can help over-sharpened pools).}
When the pool is over-sharpened ($\Delta > 0$), a weak judge may provide useful flattening; dropping it can fail to help because the pool already places too much concentrated mass on a small set of score buckets.

\paragraph{Interpretation.}
Proposition~8 reveals an asymmetry that is invisible in the mode-level veto analysis of Proposition~4. Weak judges are reliably harmful when the pool is already too flat, but can be harmless or even useful when the pool is too sharp. Sparse repair is therefore a calibration-aware intervention, with no universal preference for smaller pools.

\paragraph{Scope.}
The equal-weight analysis in Proposition~8 characterizes the fixed-pool baseline that JudgeMoE's learned gate improves upon. Its role is to identify which judges a fixed pool excludes; the learned gate can then adaptively reweight the surviving pool. Appendix Table~\ref{tab:appendix-wmt-gate-behavior} gives the matching empirical check: on WMT, the full gate already assigns negligible weight to the weak Qwen2.5 judge, but sparse retraining is still needed at 1--5 because coarse scores leave too little coherent ranking signal in the contaminated full pool.

\paragraph{Connecting the results.}
Propositions~6--8 form a coherent diagnostic chain. Proposition~6 diagnoses the pool-level calibration condition and interprets the best temperature. Proposition~7 justifies why JudgeMoE works with score distributions in addition to compressed scores, and identifies the $K \ge 3$ condition under which that advantage can exist. Proposition~8 then characterizes when removing a judge improves the pool, conditional on the calibration condition identified by Proposition~6. Together, they support the validation-guided deployment heuristic in Table~\ref{tab:deployment-summary-main}: study calibration first, keep distributional evidence whenever the score range is fine enough to support it, and use sparse repair only when the pool diagnostics indicate contamination.

\paragraph{Scope.}
Propositions~1, 3, and 4 are standard background facts stated in compact form; Propositions~6--8 are specific to the JudgeMoE formulation and serve as structured diagnostic results with limited scope. None of them imply that several small judges must beat a larger single judge. Their role is more specific: to explain why soft scores help, why temperature tuning is interpretable, why distributional fusion can add information beyond score compression, and why sparse repair helps only in particular contamination conditions.

\end{document}